\documentclass[conference]{IEEEtran}

\usepackage{tikz}
\usepackage{amsmath}

\usepackage{filecontents}

\usepackage{url}
\usepackage{hyperref}
\usepackage{graphicx}
\usepackage{tikz}
\usepackage{amsmath}
\usepackage{tabularray}
\usepackage{xspace}
\usepackage{array}
\usepackage{xcolor}
\definecolor{lightblue}{rgb}{0.55, 0.85, 0.9}

\usepackage{multirow}
\usepackage{amssymb}
\usepackage{colortbl}
\usepackage{float}
\usepackage{makecell}
\usepackage{bm}
\usepackage{caption}
\usepackage{subcaption}
\usepackage{anyfontsize}
\usepackage{cite}

\newcommand{\para}[1]{{\vspace{2pt} \bf \noindent #1 \hspace{1pt}}}

\newcommand{\genai}{{\texttt{img2img}}\xspace}

\newcommand{\eg}{e.g.,\xspace}

\newcommand{\ie}{i.e.,\xspace}

\newcolumntype{"}{!{\vrule width 1pt}}

\def\BibTeX{{\rm B\kern-.05em{\sc i\kern-.025em b}\kern-.08em
    T\kern-.1667em\lower.7ex\hbox{E}\kern-.125emX}}
\begin{document}

\title{Off-The-Shelf Image-to-Image Models Are All You Need To Defeat Image Protection Schemes}

\author{
\IEEEauthorblockN{Xavier Pleimling\IEEEauthorrefmark{1}}
\IEEEauthorblockA{
\textit{Virginia Tech}\\
Blacksburg, VA, USA\\
xavierp7@vt.edu}
\and
\IEEEauthorblockN{Sifat Muhammad Abdullah\IEEEauthorrefmark{1}}
\IEEEauthorblockA{
\textit{Virginia Tech}\\
Blacksburg, VA, USA\\
sifat@vt.edu}
\and
\IEEEauthorblockN{Gunjan Balde}
\IEEEauthorblockA{
\textit{IIT Kharagpur}\\
Kharagpur, India\\
balde.gunjan0812@kgpian.iitkgp.ac.in}
\and
\IEEEauthorblockN{Peng Gao}
\IEEEauthorblockA{
\textit{Virginia Tech}\\
Blacksburg, VA, USA\\
penggao@vt.edu}
\and
\IEEEauthorblockN{Mainack Mondal}
\IEEEauthorblockA{
\textit{IIT Kharagpur}\\
Kharagpur, India\\
mainack@cse.iitkgp.ac.in}
\and
\IEEEauthorblockN{Murtuza Jadliwala}
\IEEEauthorblockA{
\textit{University of Texas at San Antonio}\\
San Antonio, TX, USA\\
murtuza.jadliwala@utsa.edu}
\and
\IEEEauthorblockN{Bimal Viswanath}
\IEEEauthorblockA{
\textit{Virginia Tech}\\
Blacksburg, VA, USA\\
vbimal@vt.edu}
}

\maketitle

\def\thefootnote{\IEEEauthorrefmark{1}}\footnotetext{Co-lead authors.}
\def\thefootnote{\IEEEauthorrefmark{2}}\footnotetext{This work has been
accepted for publication at the IEEE Conference on Secure and
Trustworthy Machine Learning (SaTML). The final version will be
available on IEEE Xplore.}
\def\thefootnote{\arabic{footnote}}

\begin{abstract}

Advances in Generative AI (GenAI) have led to the development of various protection strategies to prevent the unauthorized use of images. These methods rely on adding imperceptible \textit{protective perturbations} to images to thwart misuse such as style mimicry or deepfake manipulations. Although previous attacks on these protections required specialized, purpose-built methods, we demonstrate that this is no longer necessary. We show that off-the-shelf image-to-image GenAI models can be repurposed as generic ``denoisers" using a simple text prompt, effectively removing a wide range of protective perturbations. Across 8 case studies spanning 6 diverse protection schemes, our general-purpose attack not only circumvents these defenses but also outperforms existing specialized attacks while preserving the image's utility for the adversary. Our findings reveal a critical and widespread vulnerability in the current landscape of image protection, indicating that many schemes provide a false sense of security. We stress the urgent need to develop robust defenses and establish that any future protection mechanism must be benchmarked against attacks from off-the-shelf GenAI models. Code is available in this repository: \url{https://github.com/mlsecviswanath/img2imgdenoiser}.

\end{abstract}


\section{Introduction}
\label{sec:intro}

\noindent The rise of Generative AI (GenAI) has heightened concerns about the unauthorized use of images, a risk that manifests itself in various forms. For instance, GenAI models are developed by indiscriminately collecting web images without obtaining permission, crediting, or compensating original creators~\cite{baack2024critical}. This technology also enables deepfakes that target individuals by altering their personal images~\cite{li2023unganable}, threatens artists' income by creating synthetic art in their style without consent, also known as a style mimicry attack~\cite{shan2023glaze}, and poses a danger to online communities through the production of harmful images~\cite{wu2025proactive}.

This has sparked significant interest in creating \textit{protection strategies} to defend against unauthorized use of images. Due to the variety of potential threats, defense measures also vary. These include strategies such as: (1) defensive watermarking~\cite{lu2024robust, gunn2024undetectable}, (2) mitigating art style mimicry~\cite{liang2023adversarial,shan2023glaze}, (3) preventing deepfake manipulations of personal images~\cite{qu2024dfrap, li2023unganable}, (4) enabling image traceability when used to customize GenAI models~\cite{li2025towards,wang2023diagnosis}, and (5) safeguarding facial privacy~\cite{le2025diffprivate, zhao2024can}. A unifying feature of all these protection strategies is the application of \textit{protective perturbations} or a \textit{protective cloak} (which are typically imperceptible) to help prevent unauthorized use. As an example, a technique known as UnGANable~\cite{li2023unganable} applies a protective cloak to an image to thwart GAN-based deepfake manipulation of that image.

Note that the defender has only one chance to protect the images. If an adversary removes this protection, there will be no way to secure the images against future misuse. Recent studies have begun to demonstrate that some of these protection methods provide a false sense of security~\cite{an2024rethinking,foerster2025lightshed,honig2024adversarial,kassis2024unmarker}. However, these attacks are generally tailored to specific types of protection strategies and require uniquely crafted AI-based methods to dismantle the protection. \textit{In this study, we demonstrate that developments in GenAI have progressed to a point where crafting specialized attacks is no longer required---off-the-shelf image-to-image (\genai) models can be easily repurposed as versatile “denoisers” to eliminate protective perturbations covering a wide variety of protection schemes.} Our key contributions are as follows:

\begin{itemize}
\item Given a protected image, we show that image translation using an off-the-shelf \genai model guided by a simple text prompt (\eg ``Denoise this image'') is sufficient to remove a variety of protective perturbations available today. We use open-source Diffusion models such as FLUX~\cite{blackfor62:online} and SD3~\cite{esser2024scaling}, and a closed-source, commercial model, GPT-4o~\cite{GPTImage1:online,hurst2024gpt}. Our approach requires no protection-specific adaptations and can be easily used by a low-skilled attacker to circumvent several protection schemes. Advances in generative processes and extensive pretraining on web-scale image datasets enable this simple attack.

\item We demonstrate the effectiveness of this simple attack using 8 case studies, covering 6 diverse protection schemes, demonstrating the versatility of the attack. In 4 of these case studies, we also compare our attack against specialized protection-specific attacks~\cite{an2024rethinking,foerster2025lightshed,honig2024adversarial,kassis2024unmarker}. Our scheme outperforms these specialized attacks.

\item Our simple approach can remove sophisticated protective perturbations. This includes perturbations to protect specialized semantic properties (\eg facial identity)~\cite{li2023unganable}, perturbations applied through the latent space~\cite{gunn2024undetectable}, and perturbations designed to survive downstream fine-tuning tasks~\cite{li2025towards}. 

\item We conduct a comprehensive user study to assess how well our denoising approach preserves utility for the attacker. Through a study of the Mist~\cite{liang2023adversarial} protection scheme, designed to prevent style mimicry attacks, we show that our denoisers produce images that are of high-quality, while preserving the expected style/content in the image. Our user study also shows that we significantly outperform recent specialized attacks against Mist, such as LightShed~\cite{foerster2025lightshed}.

\item We show that it is challenging to create a protective scheme that is resistant to our simple attack. Our efforts to integrate our denoiser into two diverse protection pipelines, \ie to create an attack-aware protection scheme, resulted in failure. The generated ``adversarial'' perturbations can still be easily removed by our denoiser. As future work, we recommend investigating robust approaches to generate protective perturbations in the low-frequency bands of an image. Our analysis of the VINE~\cite{lu2024robust} watermarking scheme shows that this is a promising approach. However, VINE's implementation of this approach still shows serious vulnerabilities against even simpler attack strategies---we find that the VINE watermark can be easily destroyed by extremely mild 0.7\% center cropping of an image.

\item Our findings highlight that future specialized protection removal attacks should invariably use off-the-shelf \genai models as a benchmark for comparison. Our approach  can even outperform specialized attacks that use a supervised learning strategy, \ie using knowledge of protected and unprotected variants of images, to remove protective perturbations~\cite{foerster2025lightshed,an2024rethinking}. 

\end{itemize}

Our results further reveal that more advanced and bigger capacity off-the-shelf \genai models are more capable at removing protections. \genai models will only continue to advance in the near future, potentially making this threat worse. We stress the need to urgently invest efforts into building robust protection schemes that can survive our attack. We have disclosed our findings to the creators of the affected protection schemes (see Section~\ref{appendix:ethics} in Appendix). 
\section{Goals, Threat Model, and Rationale}

\subsection{Goals}

\noindent We define a perturbation as any modification, typically noise, applied to an image that may be human-perceptible or not. It is usually obtained by optimizing a tailored loss function and can be added directly in pixel space or via latent space changes. Such perturbations, or \textit{protective cloaks}, are widely used as proactive defenses against unauthorized image use, including misuse of AI-generated images and copyright-protected images in training or any GenAI-based image pipeline. We show that recent GenAI advances threaten these perturbation-based defenses (\ie protective measures).

Recent breakthroughs in GenAI have led to the development of robust \textit{image translation models} or \textit{\genai} models. These models can take an existing image (source) along with a descriptive text prompt to create a new image aligned with the prompt's directive. We demonstrate that these models can be repurposed to serve as efficient \textit{denoisers}, \ie to remove any noise from the source image. We argue that this noise-removal technique is sufficient to circumvent protective mechanisms employed in current perturbation-based defense strategies. Our key research questions are as follows:

\begin{itemize}
    \item \textit{\textbf{RQ1:}} \textit{Can off-the-shelf \genai models, used as \textit{generic denoisers}, remove perturbations without any protection-specific adaptations?} We aim to minimize protection-specific customizations, \ie our denoising strategies are generic and not tailored to particular protections (\eg watermarking, deepfake manipulation protection, artwork protection). \textit{Attacker goal:} Can this process remove protective perturbations added by a defender, such as watermarks or perturbations preventing unauthorized use of images for fine-tuning GenAI models?

    \item \textit{\textbf{RQ2:}} \textit{How does our attack performance vary with increasingly capable \genai models (used as denoisers)?} \genai models continue to evolve rapidly. We experiment with a variety of open-source models in addition to an advanced commercial model. Open-source models include SD1.5~\cite{rombach2022high}, SDXL~\cite{podell2023sdxl}, SD3~\cite{esser2024scaling} and FLUX~\cite{blackfor62:online}. GPT-4o~\cite{GPTImage1:online, hurst2024gpt} is our commercial model.
    
    \item \textit{\textbf{RQ3:}} \textit{Does our denoising approach preserve the utility for the attacker?} After removing the protective perturbations, does our scheme preserve image utility for the downstream use case, \eg to use unauthorized images to train a GenAI scheme? 
    
    \item \textit{\textbf{RQ4:}} \textit{How does our simple approach compare with recent work that uses specialized perturbation-removal schemes adapted towards the protection setting?} Recall that our method is agnostic to the protection scheme and, therefore, generally applicable across many settings. Specialized schemes include methods such as UnMarker~\cite{kassis2024unmarker} designed specifically to remove watermarks from images, or INSIGHT~\cite{an2024rethinking} designed to remove protections to prevent style mimicry attacks threatening artists~\cite{shan2023glaze, liang2023adversarial}.

    \item \textit{\textbf{RQ5:}} \textit{Can our attacks, powered by simple denoisers, be neutralized by a countermeasure that leverages our denoising model in the perturbation-generation process?} Can defenders create protective perturbations that cannot be removed by our denoising schemes? 
\end{itemize} 

We address these research questions through 8 case studies, summarized in Table~\ref{tab:case-studies-overview-list}. Four case studies examine different data protection strategies, including defenses against deepfake alterations, watermarking methods, and mitigating unauthorized use of images for training GenAI models. The remaining four compare our attack with recently proposed specialized (\ie protection-specific) perturbation removal schemes.

\begin{table}[t!]
\centering
\small
\setlength{\tabcolsep}{1.2pt}
\setlength\extrarowheight{2pt}
\caption{Overview of our 8 case studies.}
\begin{tabular}{c"l|l}
\multirow{2}{*}{\textbf{\begin{tabular}[c]{@{}l@{}}Case\\ study \#\end{tabular}}} &
  \multicolumn{2}{c}{\textbf{Our attack against protection schemes}} \\ \cline{2-3}
 &
  \multicolumn{1}{c|}{\textbf{\begin{tabular}[c]{@{}c@{}}Protection type\end{tabular}}}
  &  \multicolumn{1}{c}{\textbf{\begin{tabular}[c]{@{}c@{}}Protection scheme (venue)\end{tabular}}}
  \multirow{5}{*}{\textbf{}}  \\ \Xhline{1.1pt}
1         &     \begin{tabular}[c]{@{}l@{}}Preventing deepfake\\face manipulation \end{tabular}                                     
& \begin{tabular}[c]{@{}l@{}}UnGANable (USENIX Sec'23)~\cite{li2023unganable}\\ \end{tabular}                \\
2         & \begin{tabular}[c]{@{}l@{}}In-processing\\watermarking \end{tabular}                                  
&     PRC (ICLR'25)~\cite{gunn2024undetectable}                \\
3         & \begin{tabular}[c]{@{}l@{}}Post-processing\\watermarking \end{tabular}                              
&  VINE (ICLR'25)~\cite{lu2024robust}                       \\
4         &  \begin{tabular}[c]{@{}l@{}}Traceability when \\misused for model\\personalization\end{tabular}                                   
& SIREN (IEEE S\&P'25)~\cite{li2025towards}                    \\ \cline{2-3}
\textbf{} & \multicolumn{2}{c}{\textbf{Our attack Vs. protection-specific attacks}} \\ \cline{2-3}
\textbf{} &
  \multicolumn{1}{c|}{\textbf{\begin{tabular}[c]{@{}c@{}}Protection type\end{tabular}}} 
  & \multicolumn{1}{c}{\textbf{\begin{tabular}[c]{@{}c@{}}Specialized attack (venue)\end{tabular}}} \\ \cline{2-3}
  5         & \begin{tabular}[c]{@{}l@{}}Preventing finetuning\\-based style mimicry\\  \end{tabular}                                 
  & \begin{tabular}[c]{@{}l@{}}INSIGHT (USENIX Sec'24)~\cite{an2024rethinking}\\ \end{tabular} \\         

  \begin{tabular}[c]{@{}l@{}}6\\7\\ \end{tabular}         &  \begin{tabular}[c]{@{}l@{}}Preventing Textual\\Inversion-based style\\ mimicry \end{tabular} 
  & \begin{tabular}[c]{@{}l@{}}Noisy Upscaling (ICLR'25)~\cite{honig2024adversarial}\\LightShed (USENIX Sec'25)~\cite{foerster2025lightshed} \end{tabular}                  \\

8         & \begin{tabular}[c]{@{}l@{}}Semantic\\watermarking \end{tabular}                 
& UnMarker (IEEE S\&P'25)~\cite{kassis2024unmarker}                 \\
 
\end{tabular}
\label{tab:case-studies-overview-list}
\end{table}

\subsection{Threat Model}
\noindent Since we cover a wide range of case studies, we outline a generalized threat model that is applicable to all cases. We will describe further details of the threat model pertinent to a case study in Sections~\ref{sec:attack-vs-existing-defenses} and~\ref{sec:our-attack-vs-specialized-attacks}. In this study, we adopt the perspective of an attacker, assuming access to powerful and widely available open and closed-source \genai tools.

Defenders add perturbations in latent or pixel space to prevent unauthorized image use. As attackers, we aim to remove these perturbations to enable such use, for instance by stripping watermarks or protections on personal images that prevent deepfake manipulations. We assume no knowledge of the internals or design of the protection scheme, i.e., the attack is not tailored to it. Despite this challenging setting, we show that advances in GenAI can still weaken these defenses.

\para{Countermeasures.} We also perform experiments to understand adaptive strategies by the defender based on the knowledge of our attack. This is studied in Section~\ref{sec:countermeasures}.

\subsection{Research Design and Rationale}
\label{sec:approaches-overview}

\para{Why conduct 8 different case studies and why is such a study important now?}
Table~\ref{tab:case-studies-overview-list} summarizes the 8 case studies. Examining this many cases departs from prior work on vulnerabilities of data protection schemes, which typically considers only 1–4 case studies~\cite{foerster2025lightshed, honig2024adversarial, an2024rethinking}. Our main reasons are:

(1) Our goal is to show that many perturbation-based protection schemes are vulnerable to advances in \genai technology. Our method is agnostic to the protection setting: across all 8 case studies, we use the \textbf{same} perturbation removal technique without any internal knowledge of the schemes. This shows that sophisticated, targeted mechanisms are unnecessary to uncover vulnerabilities in existing protection schemes.

(2) Perturbation-based data protection is rapidly evolving. Our survey found over 30 papers in top venues since 2024 on perturbation-based protection for images (full list in Table~\ref{tab:overview_defense_citation} in the Appendix), spanning watermarking and copyright protection~\cite{lu2024robust, gunn2024undetectable, yang2024gaussian},
art style protection~\cite{van2023anti, shan2023glaze, liang2023adversarial},
facial privacy~\cite{le2025diffprivate, zhang2025segue, zhao2024can},
deepfake manipulation~\cite{qu2024dfrap, li2023unganable, wang2022anti}, and data traceability~\cite{li2025towards, wang2023diagnosis}.  Similar efforts appear in other modalities, \eg video~\cite{low2022adverfacial, opom2022, song2024correction},
audio~\cite{meerza2024harmonycloak, zhang2023mitigating, yu2023antifake},
and text~\cite{zhang2024text, zhang2024random, salim2024impeding}.
It is infeasible to conduct a study that covers all these works. Instead, we carefully choose 8 case studies within the image modality. 

\textit{In the absence of our study, this research area will likely evolve into many disparate attempts to provide data protection without realizing that many protection pathways are easily threatened by off-the-shelf \genai schemes.}

\para{How were the 8 case studies chosen?}
We selected 8 case studies based on three criteria: \textit{(1) High Performance:} We chose protection schemes and specialized attacks that significantly outperform prior methods. Notably, 6 of the 8 works were published in 2025 in top venues, representing the state-of-the-art. \textit{(2) Diversity:} We cover a range of threats, including style mimicry, misuse of data for model personalization, watermark removal, and deepfake manipulation. \textit{(3) Reproducibility:} We prioritized methods with publicly released checkpoints, source code, and data. 

\para{What is the novelty?}We highlight two contributions. First, \textit{empirical simplicity}: we show that off-the-shelf \genai models can effortlessly remove complex, diverse protective cloaks. Second, \textit{recombinant novelty}: Beyond our empirical insights, we identify a systemic risk: foundation models act as a convergent threat vector, rendering diverse security problems susceptible to the same class of attacks.

\para{How do we navigate the complexity of evaluating 8 case studies?} Given the diverse methodologies across our 8 case studies, a unified benchmark is infeasible. Instead, we tailor metrics and datasets to each specific threat model, with the exception of Noisy Upscaling~\cite{honig2024adversarial} and LightShed~\cite{foerster2025lightshed}, which we evaluate jointly due to their shared settings. To maintain focus, we present primary findings in the main text, relegating implementation details and auxiliary results to the Appendix.

\section{Attack Methodology} 
\label{sec:denoising-methodology}
 
\noindent Given a protected image, we use an off-the-shelf \genai model, guided by a text prompt, as a ``denoiser'' to remove the (protective) perturbations in the image. There is no protection-specific adaptation or further fine-tuning of the \genai model. 

\para{Denoising models.} We use five models: four open-source Diffusion models and one closed-source commercial model.

\begin{itemize}
\item \textit{FLUX} (FLUX.1 [dev])~\cite{blackfor62:online}, a Diffusion-based model, enables photorealistic image generation and high-fidelity editing with 12B parameters.

\item \textit{SD3} (Stable Diffusion 3 Medium)~\cite{esser2024scaling} with 2B parameters is one of the latest models from Stability AI.

\item \textit{SDXL} (Stable Diffusion XL Refiner)~\cite{podell2023sdxl} with 6.6B parameters employs a two-stage ensemble model that offers superior composition and text-instruction following capabilities, just behind SD3 and FLUX in quality evaluations~\cite{esser2024scaling, Announci35:online}.

\item \textit{SD1.5} (Stable Diffusion 1.5)~\cite{rombach2022high} with approximately 890M parameters is a widely used model.

\item \textit{GPT-4o} (GPT Image 1)~\cite{GPTImage1:online, hurst2024gpt} is an Autoregressive model from OpenAI that generates extremely high-quality images~\cite{Introduc24:online, yan2025gpt}. Currently, architectural and training details are unknown. OpenAI has been consistently releasing high-performing GenAI models~\cite{OpenAIRe94:online} and we expect GPT-4o to perform significantly better than open-source models. Due to GPT-4o's pricing and our budget restrictions, we use GPT-4o on a subset of images for certain case studies. The subset is carefully chosen to be the most difficult for open-source models to denoise.

\end{itemize}

\para{Denoising methodology.} Given a perturbed image, we use a simple text prompt (e.g., ``Denoise the image'') to guide generation of the denoised image. All models support prompt-guided generation, so any user with API access and a chosen prompt can denoise images this way. Prompts can be \textit{positive} (e.g., ``Denoise the image'') to condition denoising on, or \textit{negative} (e.g., ``Add noise to the image'') to condition against. Positive and negative prompts are tokenized and used to condition attention blocks during denoising. We evaluate 8 positive–negative prompt pairs (Table~\ref{tab:combinations} in the Appendix). The FLUX model does not support negative prompts and is tested only with positive prompts. \textit{All prompts are simple, intuitive denoising instructions; we use no special rules or prompt optimizations.}

In all experiments, we input images to each denoising model at a resolution of $512 \times 512$ (the original training resolution for many models).\footnote{In Case Studies 1, 6, and 7, the initial datasets contain $256 \times 256$ images, which we upscale to $512 \times 512$ using the Stable Diffusion Upscaler~\cite{rombach2022high}.} Other hyperparameter settings for each case study are detailed in Sections~\ref{sec:attack-vs-existing-defenses} and~\ref{sec:our-attack-vs-specialized-attacks}.

\para{Why off-the-shelf \genai models?} Our intuition is based on the following key characteristics of current models:

\textit{(1) Latent space representation to compress irrelevant information:} All four Diffusion models we use operate in a latent space, \ie the source image is encoded into a lower-dimensional latent space, manipulated, and then decoded back to the pixel space. This compression captures the most perceptually relevant features, thus potentially removing any noise (perturbations) in an image.

\textit{(2) Advances in generative process aids with denoising:}  We use both Diffusion and Autoregressive (GPT-4o) models. Diffusion models learn by adding noise to the latent representation in the Forward Diffusion process, followed by reconstructing the source image by iteratively removing the noise at each step in the Reverse Diffusion process. This Diffusion process has significantly advanced over time. We select models that utilize various Diffusion processes. SD1.5 and SDXL utilize SDEdit~\cite{meng2021sdedit} for realistic image synthesis through iterative denoising with a Stochastic Differential Equation. SD3 and FLUX are Rectified Flow models that transform noise to data linearly, improving efficiency with fewer denoising steps and improving image quality. This noise-removal process of Diffusion models makes them suitable for removing perturbations.

During inference, we use two key hyperparameters: \textit{(a) Strength:} A parameter ranging from 0 to 1 that determines the amount of noise added, where 0 adds no noise and 1 dissolves the image into random noise. Higher values can more effectively destroy perturbations, but risk altering source characteristics (\eg facial identity). \textit{(b) Number of inference steps:} This parameter sets the number of denoising steps, with more steps enhancing image quality but slowing inference. 

Autoregressive models such as GPT-4o, despite unknown architecture, use a generative process aiding effective denoising. These models have strong contextual awareness~\cite{Introduc24:online}, \ie the generation of a new pixel is conditioned on the source image and all the pixels it has already generated. This can enhance the correction of corrupted (perturbed) pixels. Similarly, the U-Net backbone~\cite{ronneberger2015u} in Diffusion models uses a wide receptive field that captures contextual information surrounding a larger area, helping to differentiate between essential details and noise that should be removed.

\textit{(3) Improved knowledge of distribution of clean images:} Current models are trained on web-scale datasets (\eg LAION-5B~\cite{schuhmann2022laion}) of high quality, noise-free images, helping to better learn the mapping from noisy inputs to clean outputs.

\textit{(4) Guidance-based generation capability:} We steer generation with a text prompt that emphasizes noise removal. The strong text-conditioning abilities of these models further improve denoising. Models like GPT-4o are natively multimodal, providing a deep understanding of language–vision relationships. In Section~\ref{sec:more-analysis}, we show that prompt-guided denoising often outperforms prompt-free denoising.

\section{Our Attack Against Existing Defenses}
\label{sec:attack-vs-existing-defenses}
\noindent We evaluate our attack on data protection schemes through four case studies, addressing \textit{\textbf{RQ1}}–\textit{\textbf{RQ3}}. For each study, results use the best of eight prompt combinations by denoising performance (prompts in Table~\ref{tab:combinations}, Appendix).

\textit{Baselines:} We compare our attack with the perturbation removal baselines used in each case study. In addition, we use a common baseline based on DiffPure~\cite{nie2022diffusion}, a denoising strategy originally proposed to purify adversarial samples in classification settings. DiffPure uses a Diffusion model based on DDPM~\cite{ho2020denoising}, which operates in the pixel space, unlike our approach, which operates in the latent space (LDM~\cite{rombach2022high}). DiffPure does not require a prompt and instead relies on the Diffusion process to remove perturbations.


\subsection{Case Study 1: Mitigating Deepfake Manipulations (USENIX Security'23)}

\para{Defense background.} UnGANable~\cite{li2023unganable} protects face images (\textit{target} images) from unauthorized GAN-based deepfake manipulations. In such attacks, the adversary inverts the target image into a latent code (\textit{GAN inversion}), modifies it, and reconstructs an image with desired facial features. UnGANable adds imperceptible perturbations to the target image to disrupt inversion, yielding inaccurate latent codes and blocking manipulation. \textit{Our goal is to remove these protective perturbations to enable successful GAN inversion, verified by matching the reconstructed image’s facial identity with the target image.}

\para{Experimental setup.} Detailed setup is in Appendix~\ref{appendix:unganable-settings}.

\textit{Defense setup:} We reproduce the evaluation in UnGANable work focusing on protecting face images generated using StyleGANv2~\cite{karras2020analyzing}. 
We start with 500 $256\times256$ face images, from which we eliminate images that are not successfully protected by UnGANable. We use the black-box setting of UnGANable, \ie the defender has no knowledge of the attacker's GAN model. UnGANable bounds the perturbations using the $L_\infty$ measure using the budget parameter $\epsilon$. We experiment with $\epsilon$ values of 0.05, 0.06 and 0.07 which achieved the best results in the UnGANable work.

\textit{Baselines for comparison:} We compare our attack with Gaussian Smoothing (used by UnGANable) and DiffPure~\cite{nie2022diffusion}. 

\textit{Attack evaluation metrics}: (1) \textit{Matching Rate:} We measure the percentage of reconstructed images that match the identity of the corresponding target images, indicating a successful attack. Values range from 0 to 1, with 1 indicating a perfect attack. (2) \textit{Utility measures:} We compare the denoised images with the corresponding original target images using the same utility measures used by UnGANable. This includes Structural Similarity Index Measure (SSIM)~\cite{wang2004image}, Peak Signal-to-Noise Ratio (PSNR), and Mean Squared Error (MSE). Higher PSNR and SSIM and lower MSE indicate better attack performance.

\begin{table}[!t]
\centering
\small
\setlength{\tabcolsep}{1.5pt}
\setlength\extrarowheight{2pt}
\caption{Denoising results for UnGANable. The best results for each metric are \textbf{bolded}. The highlighted row indicates the best model of our pipeline in terms of performance and utility.}
\begin{tabular}{l"c|c|c|c}
\textbf{}                & \multicolumn{4}{c}{\textbf{$\epsilon = $ 0.06}}                                                                                                                                             \\ \cline{2-5}
\textbf{Attacker}        & \textbf{\begin{tabular}[c]{@{}c@{}}Matching\\ Rate  ↑\end{tabular}} & \textbf{PSNR ↑}   & \textbf{SSIM ↑}  & \textbf{MSE ↓}    \\ \Xhline{1.1pt}
\textbf{No attack}    & 0.00\%                                                           & 32.417          & 0.924          & 0.0006    \\
\textbf{Smoothing 
} & 63.25\%                                                           & \textbf{32.355}          & \textbf{0.947}          & \textbf{0.0006}     \\
\textbf{DiffPure}        & 48.29\%                                                           & 25.781          & 0.830          & 0.0028                  \\ \Xhline{1.1pt}
\textbf{SD1.5}      & 63.25\%                                                           & 29.077          & 0.904          & 0.0013        \\
\textbf{SDXL}   & 63.68\%                                                          & 30.726          & 0.925          & 0.0010 \\
\rowcolor[gray]{0.85} \textbf{SD3}    & \textbf{77.78\%}                                                  & 31.488 & 0.937 & 0.0007\\
\textbf{FLUX}   & 76.07\%                                                  & 31.552 & 0.941 & 0.0007
\end{tabular}
\label{tab:ungannable-small}
\end{table}

\begin{figure}[t]  
\centering
\includegraphics[width=8.75cm]{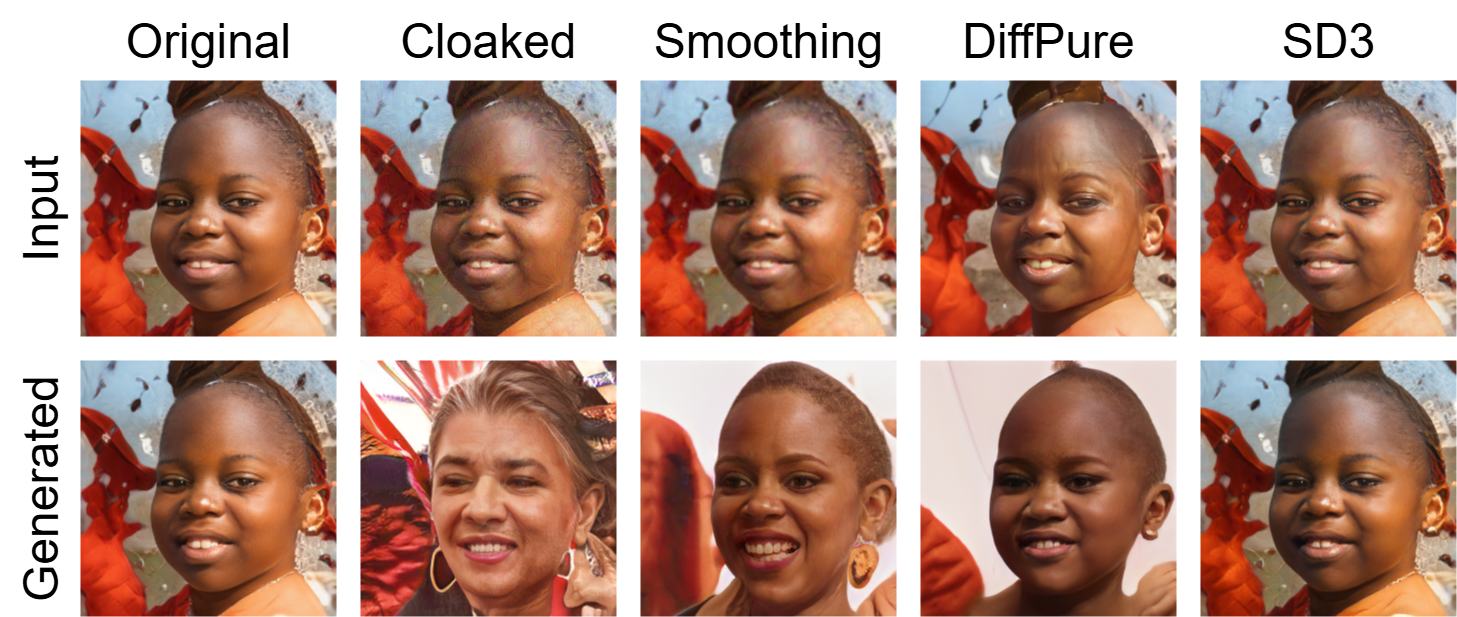} 
\caption{Images in bottom row are generated with StyleGANv2 using their respective top row image as an input. SD3 better restores the original face features compared to the baselines.}
\label{fig:unganable-images}
\vspace{-2ex}
\end{figure}

\para{Results.} We report the performance for Prompt C6 (Table~\ref{tab:combinations}) that yielded the best denoising results. Table~\ref{tab:ungannable-small} shows the results for $\epsilon=0.06$. Results for other chosen $\epsilon$ values are in Table~\ref{tab:ungannable} (Appendix) providing similar takeaways.

\textit{\textbf{RQ1:}} Our attack easily circumvents UnGANable. Without an attack, all images have a Matching Rate of 0.0\% (\ie perfect defense). Using the SD3 model, at $\epsilon=0.06$, we achieve a high Matching Rate of 77.8\%, significantly outperforming the baseline attacks (Gaussian Smoothing and DiffPure). The fact that Gaussian Smoothing can achieve a Matching Rate of 63\% further highlights the limitations of UnGANable. Note that any non-zero Matching Rate is undesired because it places those individuals at risk for serious deepfake manipulation threats.

\textit{\textbf{RQ2:}} Models with more effective diffusion/generative processes (Section~\ref{sec:denoising-methodology}) are indeed more capable. FLUX and SD3 outperforms SDXL and SD1.5. GPT-4o (not shown in Table~\ref{tab:ungannable-small}) when applied to the remaining images where SD3 failed, boosted the SD3 Matching Rate from 77.8\% to 78.6\%.

\textit{\textbf{RQ3:}} Our attack preserves data utility. We show sample images in Figure~\ref{fig:unganable-images}, and additional samples in Figure~\ref{fig:unganable-images-more} (Appendix). Our attack achieves comparable data utility measures (SSIM, PSNR and MSE) when compared to Gaussian Smoothing. We outperform DiffPure based on PSNR and SSIM, and achieve comparable and low MSE scores. 

GPT-4o substantially enhanced the image quality to the point where the denoised image is of higher quality than the target image, rendering our chosen utility metrics ineffective. We measure this using the SER-FIQ~\cite{terhorst2020ser} metric, a state-of-the-art referenceless face image quality metric. GPT-4o improved the SER-FIQ scores of the target images from 0.54 to 0.64. We further discuss this metric in Appendix~\ref{appendix:unganable-settings}. A GPT-4o image sample is shown in Figure~\ref{fig:unganable-images-gpt4o} (Appendix).

We highlight two key findings.\\
\textsc{\colorbox{lightblue}{Finding 1.}} \textit{Off-the-shelf \genai models can even remove sophisticated perturbations aimed at protecting specialized semantic properties, \ie facial identity features.}\\
\textsc{\colorbox{lightblue}{Finding 2.}} \textit{\genai models that perform translation within a latent space are more successful at stripping away protections than Diffusion-based approaches that operate directly in pixel space (DiffPure). This observation further corroborates our intuition discussed in Section~\ref{sec:denoising-methodology}.}


\subsection{Case Study 2: In-Processing Watermark (ICLR'25)}

\para{Defense background.} PRC Watermark~\cite{gunn2024undetectable} is the state-of-the-art method for in-processing watermarking in the latent space, where the watermark is applied during image generation from a prompt. PRC Watermark uses a cryptographically pseudo-random pattern that is embedded across the latent space, allowing operation at a semantic level and helping to preserve image quality. It outperforms Stable Signature \cite{fernandez2023stable} and Tree-Ring Watermarking (TRW)\cite{wen2023tree}, and is shown to be robust to pixel-level watermark removal attacks~\cite{zhao2025invisible}. 
 
\textit{Our goal is to remove the watermark so that watermark verification fails.}

\para{Experimental setup.} Detailed setup is in 
Appendix~\ref{appendix:attack-study2-settings}.

\textit{Defense setup:} We reproduce PRC Watermark's evaluation pipeline using randomly sampled prompts from the Stable Diffusion Prompt (SDP) dataset~\cite{stablediffusionprompts:online} to produce 500 $512\times512$ images with and without watermarking. 

\textit{Baselines for comparison:} Similar to the original work, we use Gaussian Smoothing and Regen-VAE~\cite{zhao2025invisible}. Regen-VAE is an attack specially designed to remove watermarks. It uses a VAE model to create a latent code that is subsequently noised and then reconstructed to disrupt the watermark. There are Regen-VAE variants, \textit{B} and \textit{C}, based on state-of-the-art image compression performance (see details in Appendix~\ref{appendix:attack-study2-settings}). We also include DiffPure.

\textit{Attack evaluation metrics:} (1) \textit{TPR@FPR:} Similar to the original work, we calculate TPR@FPR (FPR=0.00001), ranging between 0 and 1, to measure watermarking performance. Lower values indicate better attack performance. (2) \textit{Utility measures:} Similar to the original work, we use PSNR and SSIM (interpreted in the same way as Case Study 1) to measure impact on image quality. In this case, these metrics are computed between watermarked and denoised images. Additionally, we use Kernel Inception Distance (KID)~\cite{binkowski2018demystifying}, computed between unwatermarked and denoised images.\footnote{PRC Watermark and other case studies used the FID metric~\cite{heusel2017gans}, but this metric is known to show significant biases for smaller sample sizes like ours.} A lower KID value indicates better attack performance.

\begin{table}[!t]
\centering
\small
\setlength{\tabcolsep}{1.5pt}
\setlength\extrarowheight{2pt}
\caption{Denoising results for PRC Watermark. FLUX demonstrates the best balance between performance and utility.}
\begin{tabular}{l"c|c|c|c}
\textbf{Attacker}         & \textbf{TPR@FPR ↓} & \textbf{PSNR ↑} & \textbf{SSIM ↑} & \textbf{KID ↓}  \\ \Xhline{1.1pt}
\textbf{No attack}     & 1.000              & -               & -               & 0.0183          \\
\textbf{Smoothing (Avg.)} & 1.000              & \textbf{30.656}          & \textbf{0.868}           & 0.0294          \\
\textbf{DiffPure}         & 0.280              & 21.694          & 0.305           & 0.0302          \\
\textbf{Regen-VAE B}      & 0.312              & 28.859          & 0.770           & 0.0618          \\ \Xhline{1.1pt}
\textbf{SD1.5}            & 0.878              & 23.314          & 0.611           & 0.0248          \\
\textbf{SDXL}             & \textbf{0.000}     & 24.018          & 0.652           & \textbf{0.0142} \\
\textbf{SD3}              & 0.262              & 25.613 & 0.700  & 0.0196 \\
\rowcolor[gray]{0.85} \textbf{FLUX}             & 0.258     & 28.042 & 0.775  & 0.0196
\end{tabular} 
\label{tab:prc-watermark}
\end{table}

\begin{figure}[t]
\centering
\includegraphics[width=8.8cm]{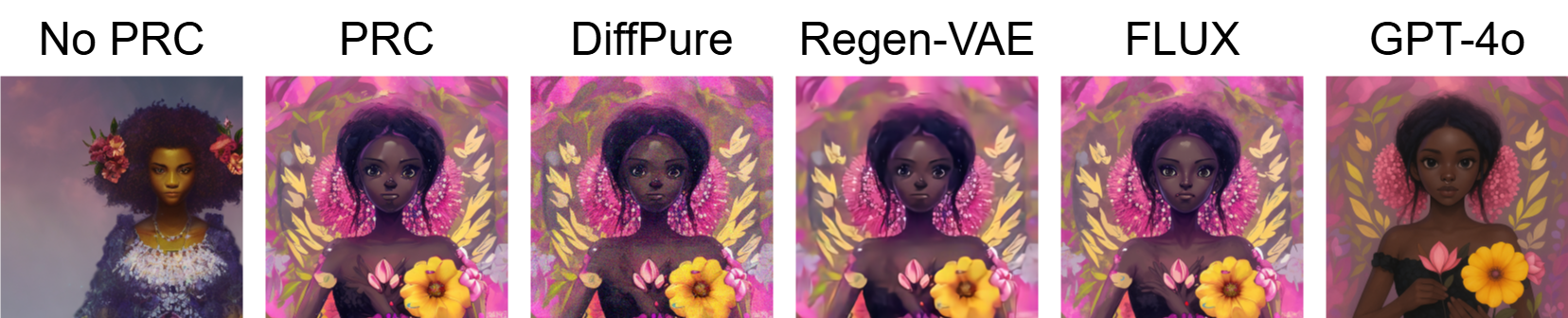}  
\caption{Qualitative image samples for PRC Watermark. Regen-VAE causes the image to appear blurrier in detail and DiffPure causes the image to appear more sharp and distorted.  FLUX is able to preserve better quality during denoising.}
\label{fig:prc-images}
\end{figure} 

\para{Results.} Prompt C8 (Table~\ref{tab:combinations}) is most effective. The key results are in Table~\ref{tab:prc-watermark}, with additional results in Table~\ref{tab:prc-watermark-full} (Appendix).

\textit{\textbf{RQ1:}} Our attack is highly effective at removing the PRC Watermark. TPR@FPR is at 1.0 when there is no attack against images, indicating perfect watermarking performance. FLUX provides the best balance in terms of reducing TPR@FPR, while maintaining image quality. FLUX significantly reduces the TPR@FPR value to 0.258 outperforming all the baselines.

\textit{\textbf{RQ2:}} Models with more advanced generative processes and larger capacity are in fact more effective, \ie FLUX, SD3, and SDXL outperform SD1.5. 
GPT-4o is applied to a subset of 100 images in which FLUX failed to remove the watermark (sampling strategy in Appendix~\ref{appendix:attack-study2-settings}). GPT-4o results are in Table~\ref{tab:prc-watermark-gpt} (Appendix). After applying GPT-4o, TPR@FPR was reduced to 0.060 when combined with the results of FLUX, further highlighting the capabilities of more advanced models.

\textit{\textbf{RQ3:}} Our attack has a minimal impact on image quality. Figure~\ref{fig:prc-images} shows sample images, and additional samples are in Figure~\ref{fig:prc-images-more} (Appendix). Using FLUX, we achieve the highest SSIM of 0.775, among attacks that degrade TPR@FPR. Among the effective baselines (where TPR@FPR is reduced), we have comparable PSNR values to Regen-VAE and outperform DiffPure. Gaussian Smoothing, despite its better utility, was completely ineffective in lowering TPR@FPR. We also achieve a low KID score of 0.0196, outperforming all baselines. Similar to Case Study 1, GPT-4o significantly enhanced image quality after denoising. Therefore, we use a referenceless widely-used metric called BRISQUE~\cite{mittal2012no} (lower values are better).\footnote{The SER-FIQ~\cite{terhorst2020ser} metric used in Case Study 1 is designed for face images and therefore not suitable here.} We justify this metric in Appendix~\ref{appendix:attack-study2-settings}. GPT-4o achieves a lower BRISQUE score of 2.859 for denoised images, compared to 5.164 for FLUX.

\textit{Finding 2} holds in this case as well. In addition:\\
\textsc{\colorbox{lightblue}{Finding 3.}} \textit{Perturbations added through latent-space manipulations can also be removed using \genai models available today.}\\
\textsc{\colorbox{lightblue}{Finding 4.}} \textit{Off-the-shelf \genai models outperform specialized watermark removal attacks (Regen-VAE).}


\subsection{Case Study 3: Post-Processing Watermark (ICLR'25)}

\para{Defense background.} VINE~\cite{lu2024robust} is a post-processing watermarking scheme, where the watermark is applied to an existing image. VINE is the state-of-the-art method in terms of robustness against image-editing techniques. VINE achieves robustness by adding invisible watermarks to the low-frequency bands of an image. The intuition is that image editing or image regeneration (like our methodology) tends to remove patterns in the high-frequency bands, and therefore embedding the watermarks in low-frequency bands can provide resilience. \textit{Our goal is to remove the watermark such that the watermark verification fails.}

\para{Experimental setup.} Detailed setup is in Appendix~\ref{appendix:attack-study3-settings}.

\textit{Defense setup:} We reproduce VINE's evaluation pipeline and apply their watermark to 1000 $512 \times 512$ images sourced from W-Bench~\cite{lu2024robust}.

\textit{Baselines for comparison:} VINE evaluates robustness against two image regeneration attacks: Stochastic Regeneration~\cite{meng2021sdedit,zhao2025invisible} and Deterministic Inversion~\cite{song2020denoising, mokady2022null}. We include these methods as baselines. Both methods use different Diffusion-based strategies to add noise to an image and then denoise it, but unlike our approach, no prompt is involved. They were not originally designed for watermark removal and were instead proposed for more general use cases (\eg image editing). Both image regeneration methods use the SD2.1 model~\cite{rombach2022high}. We also compare with Gaussian Smoothing, DiffPure, and Regen-VAE (see Case Study 2).

\textit{Evaluation metrics:} We use the same metrics used by VINE. (1) \textit{TPR@FPR:} We calculate TPR@FPR (FPR=0.001) to measure watermarking performance. (2) \textit{Utility measures:} We use PSNR, SSIM, and KID. These metrics can be interpreted similarly to Case Study 2. From VINE, we also use Learned Perceptual Image Patch Similarity (LPIPS)~\cite{zhang2018unreasonable}.
LPIPS values lie between 0 and 1 and lower values indicate better attack performance. All these metrics are calculated between unwatermarked and watermarked/denoised images, since we are considering a post-processing watermarking scheme.

\begin{table}[!t]
\centering
\small
\setlength{\tabcolsep}{1.5pt}
\setlength\extrarowheight{2pt}
\caption{Denoising results for VINE. FLUX exhibits the best balance between performance and utility.}
\begin{tabular}{l"c|c|c|c|c}
\textbf{Attacker}                & \textbf{TPR@FPR ↓} & \textbf{PSNR ↑} & \textbf{SSIM ↑} & \textbf{LPIPS ↓} & \textbf{KID ↓}  \\ \Xhline{1.1pt}
\textbf{No Denoising}            & 1.000              & 37.479          & 0.993           & 0.007            & 0.0000          \\
\textbf{DiffPure}                & 0.991              & 22.087          & 0.336           & 0.575            & 0.0033          \\
\textbf{Regen-VAE B}             & 0.976              & \textbf{29.424}          & \textbf{0.832}           & 0.275            & 0.0194          \\
\textbf{Sto. Regen.} & 0.981              & 22.325          & 0.636           & 0.270            & 0.0106          \\
\textbf{Det. Inver.} & 0.997              & 25.408         & 0.750           & 0.225            & 0.0072          \\ \Xhline{1.1pt}
\rowcolor[gray]{0.85} \textbf{FLUX}                    & \textbf{0.878}              & 26.646 & 0.775  & \textbf{0.154}   & \textbf{0.0006}
\end{tabular}
\label{tab:vine-small}
\end{table}

\begin{figure}[!t]
\centering
\includegraphics[width=8.75cm]{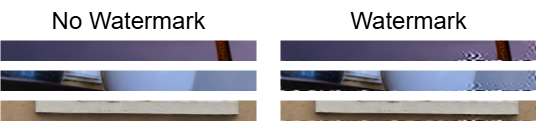}
\caption{The bottom edge of three W-Bench samples with and without VINE. VINE's watermarking creates visible perturbations on the edges of an image.}
\label{fig:vine-edges}
\end{figure}

\para{Results.} Prompt C6 (Table~\ref{tab:combinations}) is most effective. Our key results are in Table~\ref{tab:vine-small} and full results in Table~\ref{tab:vine} (Appendix).

\textbf{\textit{RQ1:}} Our attack is moderately effective at degrading TPR@FPR. Our FLUX model degrades TPR@FPR to 0.878 from 1.000, while not significantly impacting image quality. We outperform all the baseline schemes. Our performance can be attributed to the better generative process used by FLUX compared to our baselines and older Stable Diffusion schemes. 

\textit{Is VINE truly robust?} We further examined the effect of VINE restricting its perturbations to low-frequency bands, motivated by the observation that TPR@FPR did not drop substantially. 
Our analysis shows that VINE’s low-frequency design leads to perturbations that concentrate near the image boundaries and are easy to remove. As illustrated in Figure~\ref{fig:vine-edges}, these artifacts are mainly visible along the image edges. To evaluate this, we apply a simple center-cropping–based perturbation removal technique. Whereas prior center-cropping studies typically remove 10\% of the image~\cite{an2024waves}, we instead crop only 0.7\%, \ie we retain 99.3\% of the original content.
After cropping, we resize the image back to $512 \times 512$. With just 0.7\% cropping, TPR@FPR drops to 0.066, essentially erasing the watermark. Although this center-cropping method is a protection-specific adaptation (and therefore departs from our original, method-agnostic evaluation protocol), it exposes a key design weakness: VINE’s dependence on edge pixels that can be removed with minimal image modification.

\textbf{\textit{RQ2:}} More advanced and bigger capacity models are more effective. FLUX, SD3 and SDXL outperform SD1.5 based on TPR@FPR (see Table~\ref{tab:vine} in Appendix). We did not test GPT-4o as the simple center cropping attack was sufficient to fully neutralize the protection.

\textbf{\textit{RQ3:}} Our attack has minimal impact on image quality. FLUX achieves the lowest KID and LPIPS scores of 0.0006 and 0.154, respectively, outperforming all baseline schemes. For PSNR and SSIM, FLUX demonstrates values slightly below the Regen-VAE baseline, but outperforms all other baselines. Figure~\ref{fig:vine-images} (Appendix) shows sample images.

Finding 2 holds in this case as well. Additionally:\\
\textsc{\colorbox{lightblue}{Finding 5.}} \textit{Using low-frequency bands to insert the watermark is a promising idea. However, the state-of-the-art scheme (VINE) uses an implementation that produces easily removable localized perturbations in watermarked images. Future work can address this issue.}


\subsection{Case Study 4: Verifying Unauthorized Data Usage in Personalized Models (IEEE S\&P'25)}

\para{Defense background.}
SIREN~\cite{li2025towards} mitigates unauthorized use of images to fine-tune or personalize text-to-image Diffusion models. An unauthorized party can use an artist's images to personalize a Diffusion model and generate new images that mimic the artist's style, also known as \textit{style mimicry attacks}~\cite{shan2023glaze}. SIREN's idea is to enable data \textit{traceability} by adding an imperceptible perturbation or coating to an image that can be reliably learned during personalization. SIREN argues that existing watermarking schemes~\cite{shan2023glaze,van2023anti,liu2024metacloak,wang2023diagnosis,luo2023steal} are not reliable enough to transfer to output images after personalization. Watermarking schemes are independent of downstream personalization tasks and focus primarily on stealthiness. In contrast, SIREN's novel coating is designed to be relevant to the personalization task and therefore transfers to the generated images. SIREN implements a human perceptual-aware constraint using a hypersphere classification network to improve imperceptibility and traceability. \textit{Our goal is to remove the coating so that data traceability fails.} Our attack denoises the images before being used for personalization.

\para{Experimental setup.} Detailed setup is in Appendix~\ref{appendix:attack-study4-settings}.

\textit{Defense setup:}
We reproduce the experimental setup in the original work. We use the Pokemon~\cite{Pokemon:online} dataset (used by SIREN) consisting of 819 $512 \times 512$ images and associated captions generated using BLIP~\cite{li2022blip}. We use SD1.5 as the text-to-image model to personalize with ``an image of a Pokemon'' as the personalization prompt. The model is fine-tuned on the Pokemon dataset, then used to generate 1000 images. For the SIREN encoder (coating mechanism) and decoder (verification scheme), we use the provided pretrained checkpoints.

\textit{Baselines for comparison:} We compare our attack with Regen-VAE~\cite{zhao2025invisible}, which SIREN identifies as the most effective purification attack. In addition, we compare with DiffPure.

\textit{Evaluation metrics:} We use the following metrics used in the original work. (1) \textit{TPR@Significance:} We calculate TPR@Significance to measure traceability performance. Significance serves as a threshold for the Kolmogorov-Smirnov test~\cite{massey1951kolmogorov} to determine whether a generated image is traced to SIREN-coated training data.  SIREN uses a Significance of $\alpha = 10^{-9}$ where a high TPR indicates reliable data tracing. (2) \textit{Utility measures:} Similar to SIREN, we use PSNR, SSIM, and LPIPS (interpreted in the same way as Case Study 3) measured between clean (without coating) and protected/denoised images. We also use KID to measure the generation quality between clean and generated images.

\begin{table}[!t]
\centering
\small
\setlength{\tabcolsep}{1.5pt}
\setlength\extrarowheight{2pt}
\caption{Denoising results for SIREN. FLUX exhibits the best balance between performance and utility.}
\begin{tabular}{l"c|c|c|c|c}
                      \textbf{Attacker} & \textbf{TPR@Sign. ↓} & \textbf{PSNR ↑} & \textbf{SSIM ↑} & \textbf{KID ↓} & \textbf{LPIPS ↓} \\ \Xhline{1.1pt}
\textbf{No attack} & 1.000              & 39.142          & 0.880           & 0.070          & 0.016            \\
\textbf{DiffPure}     & 0.101              & 29.835          & 0.596           & 0.084          & 0.215            \\
\textbf{Regen-VAE C}  & 0.591             & \textbf{31.886}          & \textbf{0.890}           & \textbf{0.071}         & 0.085            \\ \Xhline{1.1pt}
\textbf{SD1.5}        & 0.147              & 22.348          & 0.803  & 0.100 & 0.122            \\
\textbf{SDXL}         & \textbf{0.000}     & 22.049         & 0.765           & 0.100          & 0.136            \\
\textbf{SD3}          & 0.001              & 22.935 & 0.612          & 0.112          & 0.121   \\
\rowcolor[gray]{0.85} \textbf{FLUX}         & 0.016     & 28.882 & 0.787  & \textbf{0.071} & \textbf{0.050}  
\end{tabular}
\label{tab:siren}
\end{table}

\para{Results.} Prompt C6 (Table~\ref{tab:combinations} in Appendix) yielded the best denoising performance. Results are in Table~\ref{tab:siren} (with additional results in Table~\ref{tab:siren-full} in the Appendix).

\textbf{\textit{RQ1:}} Our attack is highly effective at removing the SIREN coating. Without an attack, SIREN protection is perfect at a TPR of 1.000 allowing every image to be traced to the protection. Among our different models, FLUX performs the best in terms of TPR and utility metrics. With FLUX, TPR is reduced to 0.016. We outperform all baselines, including Regen-VAE which only achieves a TPR of 0.591. 

\textbf{\textit{RQ2:}} More advanced models with larger capacity are more effective. FLUX, SD3, and SDXL all outperform SD1.5, which still produced a low TPR of 0.147. We did not test GPT-4o as the open-source models effectively neutralize the protection.

\textbf{\textit{RQ3:}} Our attack has a minimal impact on image quality. FLUX achieves the lowest LPIPS of 0.050 outperforming all baseline schemes. Our FLUX and SD1.5 models achieve a high SSIM of 0.79 and 0.80, respectively, and is only outperformed by Regen-VAE with an SSIM of 0.89. However, recall that Regen-VAE is much less effective at lowering TPR (\ie attack performance). Our FLUX model achieves comparable KID and PSNR values to Regen-VAE and DiffPure. Figure~\ref{fig:siren-images} (Appendix) shows sample images.

In addition to Finding 2, we highlight the following:
\textsc{\colorbox{lightblue}{Finding 6.}} \textit{Protective coatings designed to survive downstream fine-tuning tasks can be effectively removed by off-the-shelf \genai models. Future work can focus on more robust data traceability schemes for personalization tasks.}


\section{Comparing Our Attack Against Protection-Specific Attacks} 
\label{sec:our-attack-vs-specialized-attacks}

\noindent We present 4 case studies that compare our attack against protection-specific attacks, covering \textit{\textbf{RQ4}}. In each case study, we aim to match or exceed the performance of existing attacks.


\subsection{Case Study 5: INSIGHT (USENIX Security'24)} 
\label{sec:case-study-5-main}
\para{Background.} INSIGHT~\cite{an2024rethinking} is an attack that removes invisible protections, enabling unauthorized use of images. It targets \textit{style mimicry attacks}, where an adversary fine-tunes a GenAI model on an artist’s images to generate new works that mimic the artist’s style. Following the original work, we attack Mist~\cite{liang2023adversarial}, a data protection scheme that adds imperceptible perturbations to images to mitigate such attacks. \textit{We compare our attack with INSIGHT, focusing on removing Mist’s protective perturbations.}

INSIGHT uses a sophisticated method to remove perturbations. Given a protected image, the key insight is to build a denoising framework that \textit{aligns} the protected image towards a carefully chosen reference image. The reference image is created by taking a photo of the protected image, which serves as a reference for how humans perceive a protected image (since perturbations are mostly imperceptible). This denoising framework includes a carefully tuned optimization scheme using VAE and UNet models.

\para{Experimental setup.} Detailed setup is in Appendix~\ref{appendix:attack-study5-settings}.

\textit{Attack setup:} We use a dataset provided by the INSIGHT work. This includes 19 $512\times512$ Mist-protected WikiArt~\cite{artgan2018} images of Van Gogh, as well as the photo-captured variants of each image (\ie reference images).\footnote{We contacted INSIGHT authors requesting a larger dataset which they were not able to provide.} The adversary aims to generate new images in the Van Gogh style by fine-tuning an SD1.5 model with DreamBooth~\cite{ruiz2023dreambooth} on the denoised images. Based on the performance in Section~\ref{sec:attack-vs-existing-defenses}, we choose FLUX as the denoising model. Configuration details are in Appendix~\ref{appendix:attack-study5-settings}. As a baseline to understand our attack performance, we study a non-adversarial setting, \ie analysis of images generated from Mist-protected images under no denoising attack.

\textit{Evaluation metrics:} (1) \textit{CLIP Accuracy:} As in INSIGHT, we use a pretrained CLIP model~\cite{radford2021learning} to assess whether generated images match the desired artistic style, reporting Top-1 and Top-3 classification accuracy. A stronger attack yields higher CLIP accuracy, \ie more images classified as the target style. (2) \textit{Utility measures:} We use the referenceless BRISQUE metric from Case Study 2 to assess generated-image quality, which is the attacker’s primary concern. Reference-based metrics (SSIM, PSNR) are unsuitable for measuring the quality of generated images, since they naturally differ from the (reference) training images. Instead, following INSIGHT, we compute PSNR and SSIM between unprotected and denoised/protected images, interpreted as in Case Study 1, and additionally report BRISQUE for denoised images.

\begin{table}[t!]
\centering
\small
\setlength{\tabcolsep}{1.5pt}
\setlength\extrarowheight{2pt}
\caption{Utility and performance results for INSIGHT and FLUX against Mist protection.  We measure CLIP accuracy to Van Gogh's style as performance. FLUX exhibits the best performance and BRISQUE utility.}
\begin{tabular}{l"cc"c"ccc}
                        & \multicolumn{2}{c"}{\textbf{\begin{tabular}[c]{@{}c@{}}CLIP \\ Accuracy\end{tabular}}}                                                  & \textbf{\begin{tabular}[c"]{@{}c@{}}Generated\\ Utility\end{tabular}} & \multicolumn{3}{c}{\textbf{\begin{tabular}[c]{@{}c@{}}Denoising\\ Utility\end{tabular}}}      \\ \cline{2-7} 
                        \textbf{Attacker} & \multicolumn{1}{c|}{\textbf{Top-1}}
                        & \textbf{Top-3}  & \textbf{BRISQUE}                                                      & \multicolumn{1}{c|}{\textbf{PSNR}}   & \multicolumn{1}{c|}{\textbf{SSIM}}  & \textbf{BRISQUE} \\ \Xhline{1.1pt}
\textbf{Mist} & \multicolumn{1}{c|}{0.0\%}          
& 1.3\%           & 60.883                                                                & \multicolumn{1}{c|}{\textbf{25.941}}          & \multicolumn{1}{c|}{\textbf{0.813}}          & 33.174           \\
\textbf{+ INSIGHT}        & \multicolumn{1}{c|}{0.7\%}            
& 48.2\%          & 29.229                                                                & \multicolumn{1}{c|}{17.945}          & \multicolumn{1}{c|}{0.393}          & 28.231           \\
\rowcolor[gray]{0.85} \textbf{+ FLUX}           & \multicolumn{1}{c|}{\textbf{4.1\%}} 
& \textbf{74.6\%} & \textbf{22.684}                                                       & \multicolumn{1}{c|}{19.570} & \multicolumn{1}{c|}{0.431} & \textbf{21.206} 
\end{tabular}
\label{tab:insight}
\end{table}

\begin{figure}[!t]
\centering
\includegraphics[width=8.75cm]{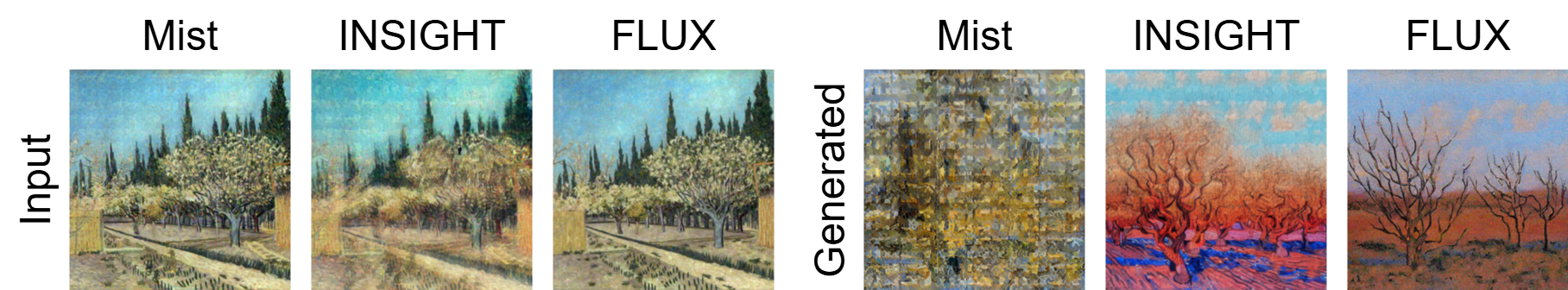}
\caption{Qualitative samples for INSIGHT.  FLUX provides the best style fit for Van Gogh, greatly improving performance compared to INSIGHT.}
\label{fig:insight-images}
\end{figure}

\para{Results.} Results are shown in Table~\ref{tab:insight}.

\textit{\textbf{RQ4:}} Our attack outperforms INSIGHT based on both Top-1 and Top-3 accuracy. Our attack achieves a high Top-3 accuracy of 74.6\%, compared to only 48.2\% for INSIGHT, rendering the Mist protection scheme ineffective.
Our attack also outperforms INSIGHT based on SSIM, PSNR, and BRISQUE scores.  We achieve the lowest BRISQUE (lower is better) of 22.68, compared to a score of 60.88 when there is no attack and 29.23 for INSIGHT. This quality is also visually reflected in the sample images of Figure~\ref{fig:insight-images} (more samples in Figure~\ref{fig:insight-images-more} (Appendix)). 

\textsc{\colorbox{lightblue}{Finding 7.}} \textit{Studies aimed at creating advanced protection-removal attacks should invariably use off-the-shelf \genai models as a benchmark for comparison. Our results underscore the advantages of employing a simpler method that leverages a robust, off-the-shelf model as opposed to a more complex attack approach.}

\textsc{\colorbox{lightblue}{Finding 8.}} \textit{Off-the-shelf \genai models can even outperform protection-removal attacks using a supervised learning approach, \eg INSIGHT uses reference (clean) images to guide the denoising process. Our attack does not require such reference images or supervision.}


\subsection{Case Studies 6 and 7: Noisy Upscaling (ICLR'25) and LightShed (USENIX Security'25)} 
\para{Background.} Noisy Upscaling~\cite{honig2024adversarial} and LightShed~\cite{foerster2025lightshed} are designed to remove protections to mitigate style mimicry. In fact, both schemes were originally evaluated against the Mist~\cite{liang2023adversarial} protection scheme (also used in Case Study 4). Therefore, we study these schemes together in this section.

\textit{Noisy Upscaling} removes perturbations by first adding Gaussian noise to an image and then applying the Stable Diffusion Upscaler~\cite{rombach2022high} to it. Directly using the SD Upscaler does not remove any noise; however, the addition of Gaussian noise in the first step helps with purification. This is because the SD Upscaler is originally trained on images augmented with Gaussian Noise. \textit{LightShed} is a more sophisticated scheme. Unlike Noisy Upscaling and our approach, LightShed uses a supervised learning scheme. As protection tools are usually made publicly available, LightShed assumes that the attacker can create a paired dataset of clean and protected versions of images. Such a dataset is used to train an Autoencoder with sophisticated loss terms to extract the perturbations in an image. The extracted perturbation is subsequently subtracted from the protected image to obtain the purified image.

\para{Experimental setup.} Detailed setup is in Appendix~\ref{appendix:attack-study6-and-7-settings}.

\textit{Attack setup:} We use the attack setup studied in the Mist work that personalizes text-to-image models for style mimicry. Given a limited set of art images (\eg 5 images), the adversary aims to generate new variations that mimic their content or style. This is implemented using Textual Inversion~\cite{gal2022image}, where a given limited set of images is used to find a new \textit{pseudo word}, $S^*$ in the embedding space of a frozen text-to-image LDM~\cite{rombach2022high} model. This pseudo word is then used to condition the LDM model by leveraging its text-to-image functionality to generate new style mimicking variations, \eg using a prompt ``a photo of $S^*$''. 

We use the LAION-Aesthetic dataset~\cite{schuhmann2022laion,laion_ev_2025} (used by LightShed) and filter 100 cat images (sampling strategy in Appendix~\ref{appendix:attack-study6-and-7-settings}). We create 20 5-image groups that are each used to optimize a pseudo-word $S^*$.  These pseudo-words are then used to conditionally generate 50 images each, obtaining a total of 1000 images.  When images are protected using Mist, the defender aims for the pseudo word estimated by the attacker to differ from ``cat,'' thereby producing images that fail to accurately represent the concept of a cat and appear more visibly distorted. In contrast, the attacker aims to generate images without any visible distortions/perturbations that accurately represent the concept of a cat.

For our attack, we choose FLUX and GPT-4o as our denoising models. We use the trained checkpoint provided by LightShed authors (for the Mist protection). More details are provided in Appendix~\ref{appendix:attack-study6-and-7-settings}. 

\begin{figure}[t]
\centering
\includegraphics[width=8.75cm]{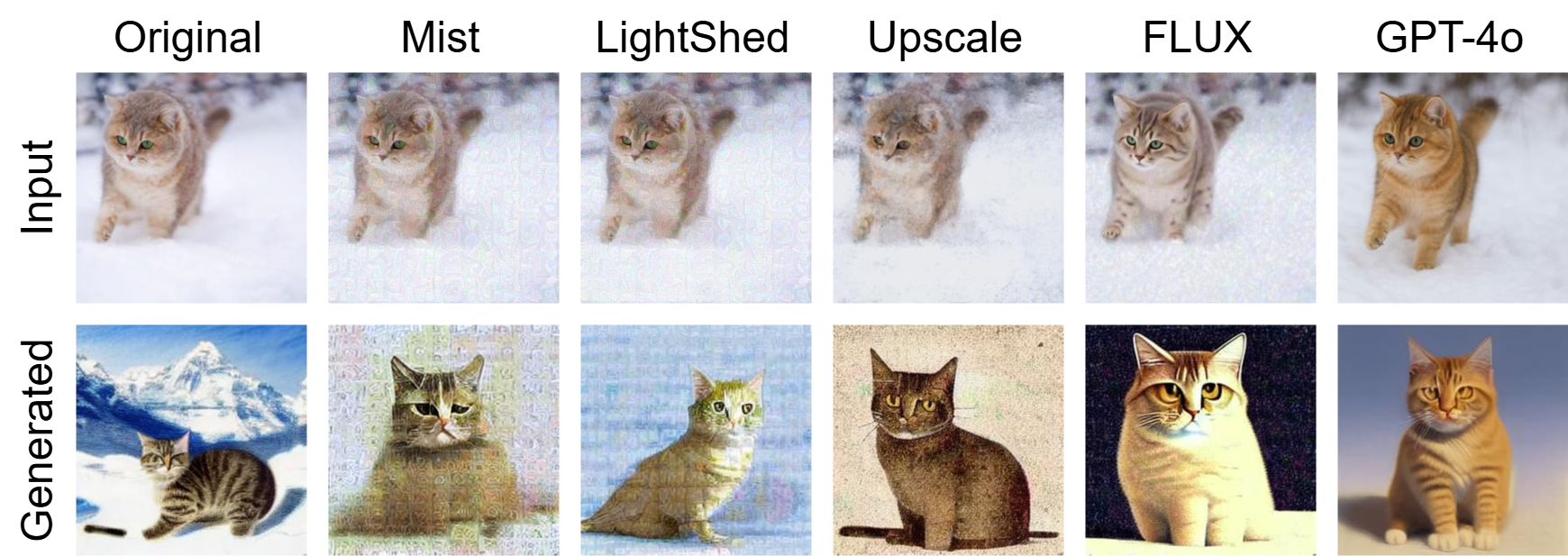} 
\caption{Images in top row are source images.  Images in bottom row are generated images from Textual Inversion. Note the high-quality sample produced when using GPT-4o.}  
\label{fig:Mist-images}
\end{figure}

\para{User study to answer RQ4.} We conduct a user study to evaluate the performance of our attack compared to LightShed and Noisy Upscaling. Note that Noisy Upscaling also relied on a user study to assess the performance of their attack. We evaluate across two dimensions: \textit{image concept-appropriateness} and \textit{image quality}. A successful attack should score highly across both dimensions. We conduct two IRB approved user studies to answer this question.

\textit{\textbf{Study setup:}} In both user studies, each participant was randomly shown 23 cat image pairs. Out of the 23, three were control image pairs (with gold standard objective answers) and rest were randomly sampled from a set of pre-generated image pairs (using the Textual Inversion pipeline). For each image pair, the users were asked to detect the better image based on image quality (noise levels, presence of artifact) and concept-appropriateness (details of the image, fit with description of cat, prompt response appropriateness and overall realism) along with attention check questions. Ultimately, every image pair is evaluated by 3 participants and we use majority voting to decide the preferred image. We also randomized the order of image pairs, added attention check questions and piloted the study to ensure data quality. More details are in Appendix~\ref{appendix:userstudy}.

\textit{In the first study}, the 20 image pairs for each participant were randomly chosen from a set of 100 image pairs. One image in the pair is always a \textit{clean} image, \ie an image generated without Mist protection (and with no attack). The other generated image varies between those produced under Mist protection, or Mist protection further denoised by Noisy Upscaling, LightShed, FLUX, GPT-4o (one chosen randomly). \textit{In the second study}, the 20 image pairs were randomly chosen from a set of 40 image pairs, where each pair contains one image generated when attacked using GPT-4o, and the other generated when attacked using Noisy Upscaling or LightShed (chosen randomly). We recruited 15 participants from Prolific academic for the first study and 6 for the second study. 

\textit{\textbf{Study results:}} Detailed results are in Table~\ref{tab:study1-input-utility-comparison} and Table~\ref{tab:study2-gpt4o-utility-comparison} (Appendix). For the first study, we compute (for each quality and concept-appropriateness question) the proportion of image-pairs where the clean image is perceived to have better utility. For the second study, we compute the proportion of image-pairs where GPT-4o images are perceived to perform better. We then leverage one-sample proportion tests (one-sided) to check if these proportions are statistically significant~\cite{ztest-proportion-test}. 
We make four important observations: 

\textit{(1)} Our participants labeled a statistically significantly greater proportion (more than 80\%) of GPT-4o images as more concept-appropriate than even clean images. In contrast, participants marked a statistically significant majority of clean images having better concept-appropriateness than LightShed images (for all concept-appropriateness metrics) and Noisy Upscaling images (for details and overall realism) images (p value ranged from $1.2\times10^{-5}$ to $0.003$). We did not observe any statistically significant differences in proportions for cases where Mist and Mist+FLUX was marked as higher/lower quality than clean image. 

\textit{(2)} Participants labeled a statistically significant majority of GPT-4o images (in 87\% to 100\% image pairs) having better concept-appropriateness than both LightShed and Noisy Upscaling images (p value ranged from $1.2\times10^{-18}$ to $1.47\times10^{-5}$). 

\textit{(3)} In terms of image quality (noise and artifact) statistically significant greater proportion of GPT-4o images are perceived to have better quality than Noisy Upscaling and LightShed (p value ranged from $2.4\times10^{-13}$ to $7.8\times10^{-8}$). In fact, for 100\% of cases users perceived GPT-4o images to have less noise compared to Noisy Upscaling and LightShed. 
\textit{Thus, our results show that GPT-4o preserves image utility to a greater extent than competing methods---in terms of both quality and concept-appropriateness.}

\textit{(4)} LightShed fails to effectively remove the Mist protection. Our participants labeled a statistically significant greater proportion of clean images (over 85\%) as more concept-appropriate (for all concept-appropriateness metrics) and of better quality (for all quality metrics) than LightShed images ($p <0.0001$). This further highlights the limitations of a supervised approach.

Findings 7 and 8 (Section~\ref{sec:case-study-5-main}) are applicable here.


\subsection{Case Study 8: UnMarker (IEEE S\&P'25)} 
\label{sec:case-study-8}

\para{Background.} UnMarker~\cite{kassis2024unmarker} is a universal watermark removal attack. It observes that robust watermarks must lie in spectral amplitudes rather than spatial structure (pixel values), \ie across different image frequency bands. UnMarker targets semantic watermarks like Tree-Ring Watermark (TRW)~\cite{wen2023tree}, which are embedded in low-frequency amplitudes and are hard to remove~\cite{zhao2025invisible,saberi2023robustness}. Note that such semantic watermarks can significantly alter content while remaining imperceptible. UnMarker's core idea is to disrupt the spectral amplitudes where robust watermarks reside, attacking both high- and low-frequency components. For low frequencies, UnMarker employs a trainable filter; additionally, it applies cropping (up to 10\%) to further weaken semantic watermarks. Recall that in Case Study 3, the VINE watermarking scheme was nearly defeated by simple cropping. \textit{We compare our attack with UnMarker, focusing on removing the TRW watermark.}

\para{Experimental setup.} Detailed setup is in Appendix~\ref{appendix:attack-study9-settings}.

\textit{Attack setup:} Following UnMarker's evaluation, we use a dataset of 100 SDP~\cite{stablediffusionprompts:online} prompts to generate watermarked and unwatermarked images. We focus on two key variants of UnMarker---a \textit{L} variant that only performs low-frequency disruption, and a \textit{HL} variant that performs both high- and low-frequency disruption. In Appendix~\ref{appendix:attack-study9-settings}, we present results for two additional variants that incorporate cropping (\textit{C}), denoted as corresponding \textit{CL} and \textit{CHL} variants.
For our attack, we choose FLUX and GPT-4o as our denoising models.

\textit{Evaluation metrics:} (1) \textit{Inverse Distance and TPR@FPR:} Following UnMarker, we use Inverse Distance and TPR@FPR for attack performance. Inverse Distance is the inverse of the average Mean Absolute Error (MAE) of extracted watermark sequences. Lower values indicate better attack performance. TPR@FPR (FPR=0.01) is also used to measure the detection performance of watermarks. Lower values are better for attacks. (2) \textit{Utility metrics:} Attacker aims to have high-quality unwatermarked images. We use the BRISQUE referenceless metric from Case Study 2 to measure the quality of the denoised images (lower is better). To further assess how much the denoised images deviate from their watermarked counterparts, we use CLIP FID~\cite{kynkaanniemi2022role}, calculated between watermarked and denoised images. Lower value would indicate that the denoised image distribution is closer to the watermarked versions.

\begin{table}[t!]
\centering
\small
\setlength{\tabcolsep}{1.5pt}
\setlength\extrarowheight{2pt}
\caption{Denoising results for TRW comparing our attack (FLUX and GPT-4o) with UnMarker on no cropping. GPT-4o exhibits best balance between performance \& utility.}
\begin{tabular}{l"cc"cc}
                        & \multicolumn{2}{c"}{\textbf{Performance}}                                                                     & \multicolumn{2}{c}{\textbf{Denoising Utility}}                                              \\ \cline{2-5} 
                        \textbf{Protection} & \multicolumn{1}{c|}{\textbf{\begin{tabular}[c]{@{}c@{}}Inv.\\ Dist. ↓\end{tabular}}} & \textbf{\begin{tabular}[c]{@{}c@{}}TPR@FPR\\ = 0.01 ↓\end{tabular}} 
                        & \multicolumn{1}{c|}{\textbf{\begin{tabular}[c]{@{}c@{}}CLIP\\ FID ↓\end{tabular}}}
                        & \textbf{BRISQUE ↓} \\ \Xhline{1.1pt}
\textbf{UnMarker (HL)}  & \multicolumn{1}{c|}{0.0162}              & 0.9011                            
& \multicolumn{1}{c|}{6.774}                     
& 62.969 \\
\textbf{UnMarker (L)}   & \multicolumn{1}{c|}{0.0166}              & 0.9560                                                             
& \multicolumn{1}{c|}{\textbf{5.771}}             
& \textbf{7.728} \\
\textbf{FLUX}           & \multicolumn{1}{c|}{0.0153}              & 0.7912                                                                     
& \multicolumn{1}{c|}{11.549}                      
& 13.958 \\
\rowcolor[gray]{0.85} \textbf{GPT-4o}         & \multicolumn{1}{c|}{\textbf{0.0149}}     & \textbf{0.6813}                                                             
& \multicolumn{1}{c|}{12.559}            
& 8.002

\end{tabular}
\label{tab:unmarker-small}
\end{table}

\begin{figure}[t]
\centering
\includegraphics[width=8.75cm]{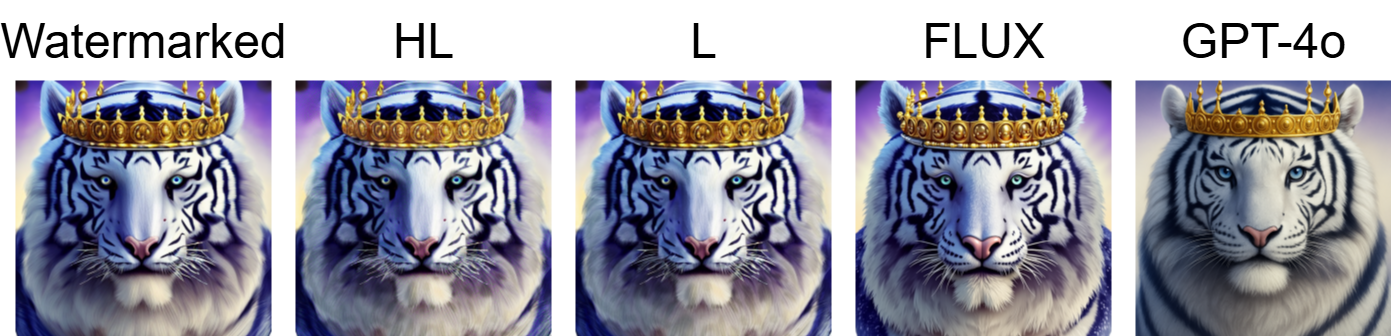}
\caption{Qualitative examples comparing UnMarker with FLUX and GPT-4o on TRW.  Note that GPT-4o provided a high quality sample that appears different from the watermarked sample but is still of the same concept.}
\label{fig:unmarker-images}
\end{figure}

\para{Results.} Results are shown in Table~\ref{tab:unmarker-small} and additional results are in Table~\ref{tab:unmarker-trw} (Appendix).

\textbf{\textit{RQ4:}} GPT-4o and FLUX outperform both variants of UnMarker (HL and L) in Inverse Distance and TPR@FPR metrics. GPT-4o reduces the TPR@FPR from 1.0 to 0.68, which is achieved while maintaining high image quality. GPT-4o achieves a low BRISQUE score, comparable to UnMarker (L). Figure~\ref{fig:unmarker-images} shows image samples. More images are in Figure~\ref{fig:unmarker-images-more} (Appendix). Figure~\ref{fig:unmarker-images} also indicates that the GPT-4o image looks slightly different from the watermarked image, but maintains the content/theme, \ie a white tiger wearing a crown with a certain color background. This further explains our CLIP FID results. Although UnMarker shows the least deviation from the watermarked version (at the cost of reduced attack performance), GPT-4o removes watermarks more effectively while regenerating similar images (with some changes).

We also experiment with UnMarker variants using 10\% cropping (CHL and CL). The results in Table~\ref{tab:unmarker-trw} (Appendix) show that UnMarker benefits significantly from cropping, outperforming our attack with a TPR@FPR of 0.18 compared to our GPT-4o's 0.56. Cropping also improves our attack. UnMarker does not adequately explain why cropping aids in semantic watermark removal. In fact, cropping is not part of their main method and is added only, as an additional step in their experimental sections. We believe, like with VINE (Case Study 3), perturbations are concentrated at image edges.

\textsc{\colorbox{lightblue}{Finding 9.}} \textit{Off-the-shelf \genai models, without any protection-specific optimizations, outperform sophisticated watermark removal attacks. However, if simple cropping is used as an additional step, protection-specific schemes like UnMarker will outperform off-the-shelf \genai approaches. This can be attributed to the spatial biases exhibited by certain watermarking schemes (such as TRW).}

\section{Countermeasures}
\label{sec:countermeasures}

\noindent We study an \textit{attack-aware defender} who aims to produce perturbations that are resilient to our removal attacks and answers \textbf{RQ5}. The key idea is to use our denoiser in the protection-generation pipeline. We conduct this study for UnGANable (Case Study 1) and SIREN (Case Study 4).\footnote{For VINE (Case Study 3), we already show that denoisers are not needed to remove protection, just simple cropping will suffice. PRC Watermark (Case Study 2) does not provide training code to conduct such an experiment.}

\para{Attack-aware UnGANable.} UnGANable's protective cloak is computed over a set number of iterations. We design a countermeasure by integrating our SDXL denoiser into this pipeline.\footnote{Other denoising models were not compatible with the older libraries used by UnGANable.} After each iteration to optimize the cloak, we denoise the adversarial image using SDXL, thereby having the next iteration account for this adversarial modification. We use 17 images where we succeeded in our denoising attack. The defender now aims to create new perturbations for these images that are resilient to our denoising attack. Appendix~\ref{appendix:unganable-countermeasure} provides more details of UnGANable's optimization objectives and our experimental configuration.

\begin{figure}[t]
\centering
\includegraphics[scale=0.17, angle=270]{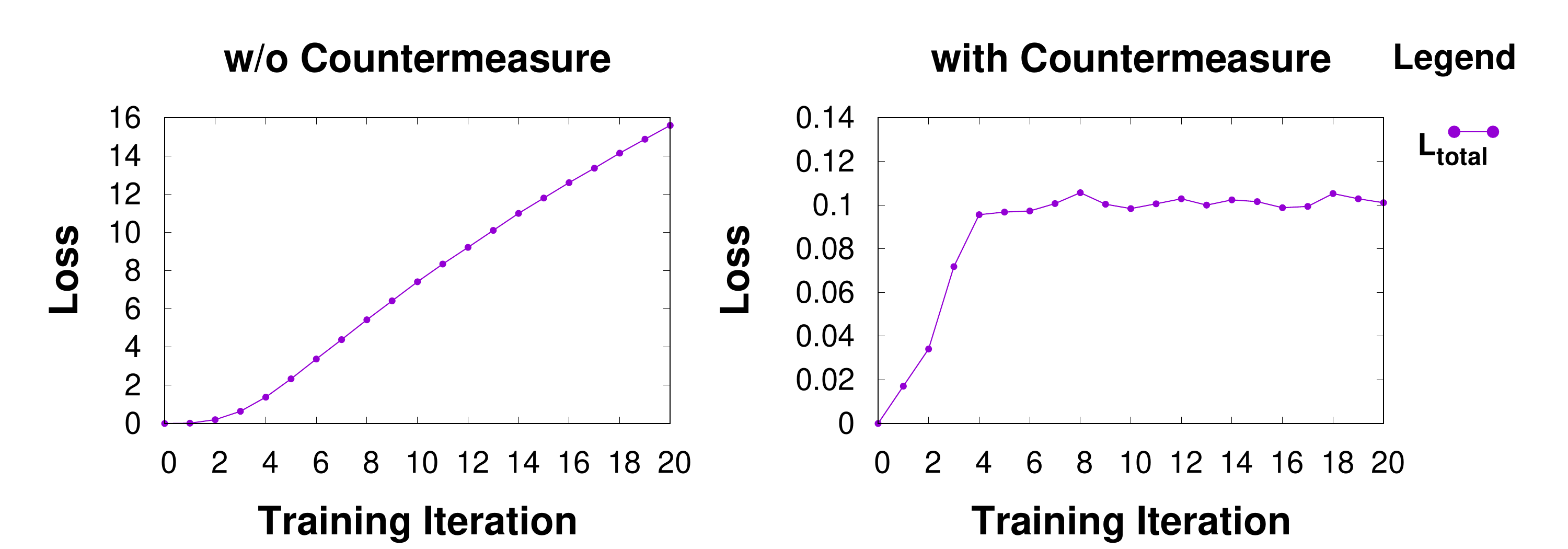} 
\caption{Loss curve for UnGANable with and without the countermeasure. Without the countermeasure (left), loss values display an increasing trend with high magnitude. With the countermeasure (right), loss values plateau after only four training iterations, exhibiting values of low magnitude. }
\label{fig:unganable-countermeasure}
\end{figure}

We find that this countermeasure is not highly effective in producing attack-resistant perturbations. The Matching Rate dropped to 83\% from 100\%, \ie only 17\% of the images survived our denoising attack. Results are in Table~\ref{tab:unganable-countermeasure} (Appendix). To understand this outcome, we analyze the UnGANable optimization loss curves in Figure~\ref{fig:unganable-countermeasure}. Note that UnGANable aims to maximize $\mathcal{L}_{total}$ (see Equation~\ref{eq:cloakv1-2} in Appendix). Without the countermeasure, the total loss steadily improves with each iteration. When the countermeasure is applied, the loss value barely exceeds 0.11 and does not improve subsequently. This change in learning behavior suggests that integrating our denoiser destabilizes UnGANable's optimization scheme.

\para{Attack-aware SIREN.} We follow a similar strategy for SIREN, by integrating our denoiser into SIREN's perturbation generation pipeline. Due to space limitations, we discuss this in Appendix~\ref{appendix:siren-countermeasure}. We show that our denoising attack can still effectively reduce TPR@FPR from 0.99 to 0.00 for samples using attack-aware perturbations. Our analysis of the loss curves again highlights a trend similar to that observed in the UnGANable countermeasure study.

\textsc{\colorbox{lightblue}{Finding 10.}} \textit{It is challenging to create protective perturbations that are resilient to off-the-shelf \genai models. Future work can study more robust protection schemes. Finding 5 further provides a promising direction.}

\section{Impact of Hyperparameters and Denoising Adaptations}\label{sec:more-analysis}

\noindent We describe ablation studies to test the impact of our attack hyperparameters, and also evaluate two adaptations to our attack: \textit{no-prompt denoising} and \textit{supervised denoising}. 

\para{Impact of hyperparameters.} We discuss the impact of two key hyperparameters of our attack: \textit{(1) Text prompt:} We use the UnGANable case study to demonstrate the impact of text prompt variations when using our SD3 denoiser. Results are in Table~\ref{tab:prc-ablation-prompt} (Appendix). At $\epsilon=0.06$, the Matching Rate (higher is better) varies from 63\% to 78\% (average of 69\%), suggesting that certain prompts can indeed offer a boost in attack performance. The standard deviation of the Matching Rate is only 4.52\%. Prompts C6 and C8 were our best prompts across all the case studies. \textit{(2) Strength parameter:} This parameter controls the amount of noise added during the Forward Diffusion process and is only applicable to our Diffusion models. We use PRC Watermark (Case Study 2) and SIREN (Case Study 4) to evaluate the impact of the strength parameter. Tables~\ref{tab:prc-ablation-str} and Table~\ref{tab:siren-ablation-str} in Appendix show the results for PRC and SIREN, respectively. As expected, we see that increasing strength does improve attack performance (\eg reduce TPR@FPR), but at the cost of reduced image utility (\eg image quality metrics).

\para{Denoising without prompts.} Our open-source denoisers (Diffusion models) can be used without a text prompt, by setting the prompt field to an empty string and the \textit{guidance scale}~\cite{Imagetoi68:online} parameter to 1. The results are shown in Tables~\ref{tab:prc-watermark-nonprompt},~\ref{tab:vine-nonprompt},~\ref{tab:siren-noprompt}, and~\ref{tab:ungannable-nonprompt} (all in Appendix) for the 4 case studies in Section~\ref{sec:attack-vs-existing-defenses}. We find that prompt-based denoising outperforms the no-prompt approach across all four case studies based on the key evaluation metrics in each case study.  One such significant improvement is the TPR@FPR for FLUX on PRC Watermark improving from 0.420 for no-prompt to a lower 0.258 with the prompt-based approach. More results are in Appendix~\ref{appendix:section8-2}.

\para{Supervised denoising.} Protection tools are usually made publicly available. This allows an attacker to create a paired dataset of protected and unprotected images. We use such a dataset to fine-tune our \genai pipeline for UnGANable (Case Study 1). We use the SDXL denoiser for this experiment. More details are in Appendix~\ref{appendix:section8-3}. With supervised denoising, we obtain a Matching Rate of 69.66\% which outperforms unsupervised SDXL's 63.68\%. However, the supervised approach does not outperform our best unsupervised denoiser---SD3 which achieves a Matching Rate of 77.78\%. This suggests that models with larger capacities and more effective generative processes can still outperform supervised approaches that use less capable \genai models. Detailed results are in Table~\ref{tab:unganable-supervised} (Appendix).

\section{Conclusion}
\label{sec:discussion}

\noindent This work demonstrates that the arms race between image protection and misuse has reached a critical turning point. Our research reveals that the prevailing assumption that removing protective perturbations requires specialized, purpose-built attacks---is now obsolete. The very GenAI technologies that necessitated these protections have evolved into powerful and universally effective tools for dismantling them. Our key finding is that off-the-shelf \genai models can be easily repurposed as ``denoisers'' to strip a wide array of protective cloaks from images using simple text prompts. Our attack not only preserves the utility of the image for an adversary but also outperforms existing specialized protection-specific attacks. As GenAI models continue to grow in capability, this threat will only become more severe. Therefore, we stress the urgent need for the research community to develop a new generation of robust protection schemes. We posit that resilience against this simple denoising attack must serve as a fundamental benchmark for any future defense mechanism.

\section*{Acknowledgments}

\noindent This material is based upon work supported by the National Science Foundation under grant award numbers 2231002, 1943351, and 2442171. This work is also partially supported by a Google Academic Research Award (GARA).

\section*{LLM Usage Considerations}
\label{sec:llm-usage}

\para{Originality.} LLMs were used for editorial purposes in this manuscript, and all outputs were inspected by the authors to ensure accuracy and originality.

\para{Transparency.} 
We ensure full reproducibility for experiments involving all local models and deterministic algorithms. However, experiments utilizing the GPT-4o API as a denoiser may exhibit minor variations over time due to the continuous updates. Experiments were conducted using publicly available datasets and code repositories, and full methodological details are provided in the paper. We released all the relevant code, datasets, and model checkpoints together with comprehensive instructions and scripts to reproduce all analyses and figures.

\para{Responsibility.} 
We performed an extensive evaluation across 8 different case studies. To prioritize efficiency and reduce environmental impact, we intentionally selected representative experimental pipelines for each of the papers instead of exhaustively replicating every reported result. For our GPT-4o experiments, API usage was intentionally minimized and targeted to the most challenging samples to establish model capabilities while containing compute and monetary costs. 
Experimental settings, selection rationale, and implementation details are fully described in the manuscript. 
All experiments were conducted on NVIDIA hardware: A100 GPUs (80 GB VRAM), Quadro RTX 8000 GPUs, Titan RTX GPUs, and A40 GPUs (48 GB VRAM).

\bibliographystyle{plain}
\bibliography{main.bib}

\appendix

\label{sec:appendix} 

\subsection{Ethical Considerations}\label{appendix:ethics}

\noindent All experiments were conducted in controlled laboratory settings using publicly available code and datasets; no deployed real-world models were attacked in the process. We also followed responsible disclosure prior to submission and communicated our findings to the authors of the protection schemes we found vulnerable: UnGANable, PRC Watermark, VINE, SIREN, Mist, and TRW. We have made code and data publicly available after verifying that their release poses no safety concerns. This supports transparency and reproducibility, and will facilitate future research focused on building stronger protections against unauthorized image use. The total cost of image editing using OpenAI's GPT-4o amounted to approximately \$73.

We place a high value on ethical human subjects research ---as a result our survey study protocol went through multiple iterations and scrutiny. Researchers and personnel from the Institutional Review Board (IRB)—referred to as the Institute Ethics Committee (IEC) in the university---were actively involved in the process. In the final protocol, participants were provided with a clear explanation of the study’s objectives, the expected time commitment, and the compensation. We also mentioned that this study involved labeling of cat images, in case they have ailurophobia. They were explicitly informed that we would not store any personally identifiable information. They could abort the study at any time. Finally, we removed identifying information like prolific ids from the final stored data and keep the data in a password protected computer situated within our university. In summary, we followed the  best ethical research practices to obtain informed consent from participants, preserve participant anonymity and ensure security of the survey data.

\begin{table*}[t!]
\centering
\normalsize
\setlength{\tabcolsep}{1.50pt}
\setlength\extrarowheight{1pt}
\caption{Published papers using perturbation-based techniques for proactive defense or protection. The publication venues are highlighted in parentheses after each group of citations.}
\begin{tabular}{l"c|l}
\textbf{Domain}                 & \textbf{Category}                                                                      & \textbf{Published Papers} \\ \Xhline{1.1pt}
\multirow{10}{*}{\textbf{Image}} & \textbf{\begin{tabular}[c]{@{}c@{}}Watermarking /\\ Copyright Protection\end{tabular}} & \begin{tabular}[l]{@{}l@{}}~\cite{fang2023flow} \textbf{(AAAI)} ~\cite{xiong2023flexible,ma2022towards,fang2022pimog} \textbf{(ACM MM)} ~\cite{zhang2024editguard,yang2024gaussian} \textbf{(CVPR)} ~\cite{ci2024ringid} \textbf{(ECCV)}\\ ~\cite{tan2024waterdiff} \textbf{(ICASSP)} ~\cite{fernandez2023stable} \textbf{(ICCV)} ~\cite{gunn2024undetectable,lu2024robust} \textbf{(ICLR)}  ~\cite{feng2024aqualora} \textbf{(ICML)}\\ ~\cite{zhang2024attack,guo2024freqmark,xian2024raw,huang2024robin,wen2023tree} \textbf{(NeurIPS)} ~\cite{luo2024wformer} \textbf{(IEEE TETCI)}\\~\cite{qiao2024scalable} \textbf{(IEEE TIFS)} ~\cite{zhang2025markplugger,meng2025latent} \textbf{(IEEE TMM)} ~\cite{lukas2023ptw} \textbf{(USENIX Security)}\end{tabular}                          \\ \cline{2-3}
                                & \textbf{Art Style Protection}                                                          & \begin{tabular}[l]{@{}l@{}}~\cite{van2023anti} \textbf{(ICCV)} ~\cite{xue2024effective} \textbf{(ICLR)}  ~\cite{liang2023adversarial,salman2023raising} \textbf{(ICML)} 
                                                              ~\cite{shan2023glaze} \textbf{(USENIX Security)} \end{tabular}        \\ \cline{2-3}
                                & \textbf{Facial Privacy}                                                              & 
                                \begin{tabular}[l]{@{}l@{}} ~\cite{sun2024diffam,wang2024simac,shamshad2023clip2protect,wang2023privacy,hu2022protecting} \textbf{(CVPR)} ~\cite{zhang2025segue} \textbf{(ICASSP)}  ~\cite{wen2023divide,yang2021towards} \textbf{(ICCV)} \\
                                ~\cite{an2024sd4privacy} \textbf{(IEEE ICME)} ~\cite{wan2024prompt} \textbf{(NeurIPS)} ~\cite{le2025diffprivate,le2024styleadv,chow2024diversity} \textbf{(PETS)} \\~\cite{wang2025beyond} \textbf{(IEEE TIFS)} ~\cite{shan2020fawkes} \textbf{(USENIX Security)} \end{tabular}                           \\ \cline{2-3}
                                & \textbf{Deepfake Manipulation}                                                         &  
                                \begin{tabular}[l]{@{}l@{}} ~\cite{huang2021initiative} \textbf{(AAAI)} ~\cite{wang2022deepfake, chen2021magdr} \textbf{(CVPR)} ~\cite{fan2024synthesizing} \textbf{(ICASSP)}
                                 ~\cite{wang2022anti} \textbf{(IJCAI-ECAI)} \\ ~\cite{dong2023restricted,qu2024dfrap} \textbf{(IEEE TIFS)} ~\cite{li2023unganable} \textbf{(USENIX Security)} ~\cite{yeh2020disrupting} \textbf{(WACV)} \end{tabular}                       \\ \cline{2-3}
                                & \textbf{Data Traceability}                                                             & ~\cite{kim2024wouaf} \textbf{(CVPR)} 
                                ~\cite{wang2023diagnosis} \textbf{(ICLR)} ~\cite{li2025towards} \textbf{(IEEE S\&P)}                          \\ \hline
\textbf{Video}                  & -                                                                                      &  \begin{tabular}[l]{@{}l@{}}~\cite{low2022adverfacial} \textbf{(ICASSP)} ~\cite{rosberg2023fiva} \textbf{(ICCV Workshop)} ~\cite{gupta2024visual} \textbf{(IEEE Access)}
~\cite{anand2020adversarial} \textbf{(IEEE ICMLA)}\\
~\cite{sun2022privacy} \textbf{(IEEE LSP)} ~\cite{lee2023defending} \textbf{(IEEE TDSC)} ~\cite{chen2022pulseedit} \textbf{(IEEE TIFS)}
~\cite{lo2021defending} \textbf{(IEEE TIP)}\\
~\cite{opom2022} \textbf{(IEEE TPAMI)} ~\cite{song2024correction} \textbf{(USENIX Security)} \end{tabular}                        \\ \hline
\textbf{Audio}                  & -                                                                                      &  \begin{tabular}[l]{@{}l@{}}
~\cite{yu2024towards, yu2023antifake} \textbf{(ACM CCS)} ~\cite{liu2023protecting} \textbf{(ACSAC)} ~\cite{wu2021adversarial,wu2020defense} \textbf{(ICASSP)}\\ ~\cite{zhang2023mitigating} \textbf{(LAMPS)} ~\cite{dong2024active} \textbf{(IEEE LSP)} ~\cite{meerza2024harmonycloak} \textbf{(IEEE S\&P)} ~\cite{cheng2023uniap} \textbf{(IEEE TDSC)}\\ ~\cite{hussain2021waveguard} \textbf{(USENIX Security)} ~\cite{wang2023vsmask} \textbf{(WiSec)}
\end{tabular}                       \\ \hline
\textbf{Text}                   & -                                                                                      & \begin{tabular}[l]{@{}l@{}}
~\cite{zhang2022improving} \textbf{(ACL)} ~\cite{christ2023undetectable} \textbf{(COLT)} ~\cite{zhang2024random} \textbf{(EACL)} ~\cite{salim2024impeding, han2020adversarial} \textbf{(EMNLP)} \\ ~\cite{zhang2024text} \textbf{(IEEE S\&P)} ~\cite{wang2022rethinking} \textbf{(IEEE TASLP)}
\end{tabular}                          
\end{tabular}
\label{tab:overview_defense_citation}
\end{table*}

\begin{table}[t!]
\centering
\small
\setlength{\tabcolsep}{1.5pt}
\setlength\extrarowheight{1pt}
\caption{Labels for \textit{prompts} and \textit{negative prompts} combinations numbered C1-C8. These prompts are used as part of our denoising pipeline to attack perturbed images. Note that FLUX does not use negative prompts.}
\begin{tabular}{l"c|c}
\textbf{Label} & \textbf{\textit{prompts}} & \textbf{\textit{negative prompts}} \\ \Xhline{1.1pt}
C1 & Denoise the image & Add noise to the image \\ \hline
C2 & Smoothen the image & Add noise to the image \\ \hline
C3 & \begin{tabular}[c]{@{}c@{}}Denoise the image while\\ preserving content of the image\end{tabular} & Add noise to the image \\ \hline
C4 & \begin{tabular}[c]{@{}c@{}}Remove adversarial\\ perturbations\end{tabular} & Add noise to the image \\ \hline
C5 & Denoise the image & \begin{tabular}[c]{@{}c@{}}Add adversarial\\ perturbation to the image\end{tabular} \\ \hline
C6 & Smoothen the image & \begin{tabular}[c]{@{}c@{}}Add adversarial\\ perturbation to the image\end{tabular} \\ \hline
C7 & \begin{tabular}[c]{@{}c@{}}Denoise the image while\\ preserving content of the image\end{tabular} & \begin{tabular}[c]{@{}c@{}}Add adversarial\\ perturbation to the image\end{tabular} \\ \hline
C8 & \begin{tabular}[c]{@{}c@{}}Remove adversarial\\ perturbations\end{tabular} & \begin{tabular}[c]{@{}c@{}}Add adversarial\\ perturbation to the image\end{tabular}
\end{tabular}%
\label{tab:combinations}
\end{table}

\subsection{Case Study 1: Mitigating Deepfake Manipulations (USENIX Security'23)}
\label{appendix:unganable-settings}

\begin{table*}[!t]
\centering
\small
\setlength{\tabcolsep}{1.5pt}
\setlength\extrarowheight{2pt}
\caption{Full denoising results for UnGANable on the best prompt (C6) highlighting its performance on perturbation budgets $\epsilon=\{0.05,0.06,0.07\}$.}
\begin{tabular}{l"c|c|c|c"c|c|c|c"c|c|c|c}
\textbf{}                & \multicolumn{4}{c"}{\textbf{$\epsilon = $ 0.05}}                                                                                & \multicolumn{4}{c"}{\textbf{$\epsilon = $ 0.06}}                                                                                & \multicolumn{4}{c}{\textbf{$\epsilon = $ 0.07}}                                                                                \\ \cline{2-13}
\textbf{Attacker}        & \textbf{\begin{tabular}[c]{@{}c@{}}Matching\\ Rate ↑\end{tabular}} & \textbf{PSNR ↑}   & \textbf{SSIM ↑}  & \textbf{MSE ↓}    & \textbf{\begin{tabular}[c]{@{}c@{}}Matching\\ Rate ↑\end{tabular}} & \textbf{PSNR ↑}   & \textbf{SSIM ↑}  & \textbf{MSE ↓}    & \textbf{\begin{tabular}[c]{@{}c@{}}Matching\\ Rate  ↑\end{tabular}} & \textbf{PSNR ↑}   & \textbf{SSIM ↑}  & \textbf{MSE ↓}    \\ \Xhline{1.1pt}
\textbf{No Denoising}    & 0.00\%                                                           & 33.880          & 0.944          & 0.0004          & 0.00\%                                                           & 32.417          & 0.924          & 0.0006          & 0.00\%                                                            & 31.218          & 0.903          & 0.0008          \\
\textbf{Smoothing 
} & 57.44\%                                                           & \textbf{32.881}          & \textbf{0.954}          & \textbf{0.0006}          & 63.25\%                                                           & \textbf{32.355}          & \textbf{0.947}          & \textbf{0.0006}          & 56.52\%                                                           & \textbf{31.856}          & \textbf{0.940}          & \textbf{0.0007}          \\
\textbf{DiffPure}        & 45.13\%                                                           & 25.810          & 0.831          & 0.0028          & 48.29\%                                                           & 25.781          & 0.830          & 0.0028          & 46.01\%                                                           & 25.731          & 0.828          & 0.0028          \\ \Xhline{1.1pt}
\textbf{SD1.5}           & 64.62\%                                                           & 29.400          & 0.913          & 0.0014          & 63.25\%                                                           & 29.077          & 0.904          & 0.0013          & 54.71\%                                                           & 28.684          & 0.894          & 0.0014          \\
\textbf{SDXL}            & 72.31\%                                                           & 31.097          & 0.932          & 0.0008 & 63.68\%                                                           & 30.726          & 0.925          & 0.0010 & 64.49\%                                                           & 30.388          & 0.918          & 0.0010 \\
\rowcolor[gray]{0.85} \textbf{SD3}             & 75.38\%                                                  & 31.908 & 0.943 & 0.0007 & \textbf{77.78\%}                                                  & 31.488 & 0.937 & 0.0007 & 71.01\%                                                  & 31.029 & 0.929 & 0.0008 \\
\textbf{FLUX}            & \textbf{76.41\%}                                                  & 31.940 & 0.947 & 0.0007 & 76.07\%                                                  & 31.552 & 0.941 & 0.0007 & \textbf{72.46\%}                                                  & 31.148 & 0.934 & 0.0008
\end{tabular}
\label{tab:ungannable}

\end{table*}

\begin{figure}[t]  
\centering 
\includegraphics[width=8.75cm]{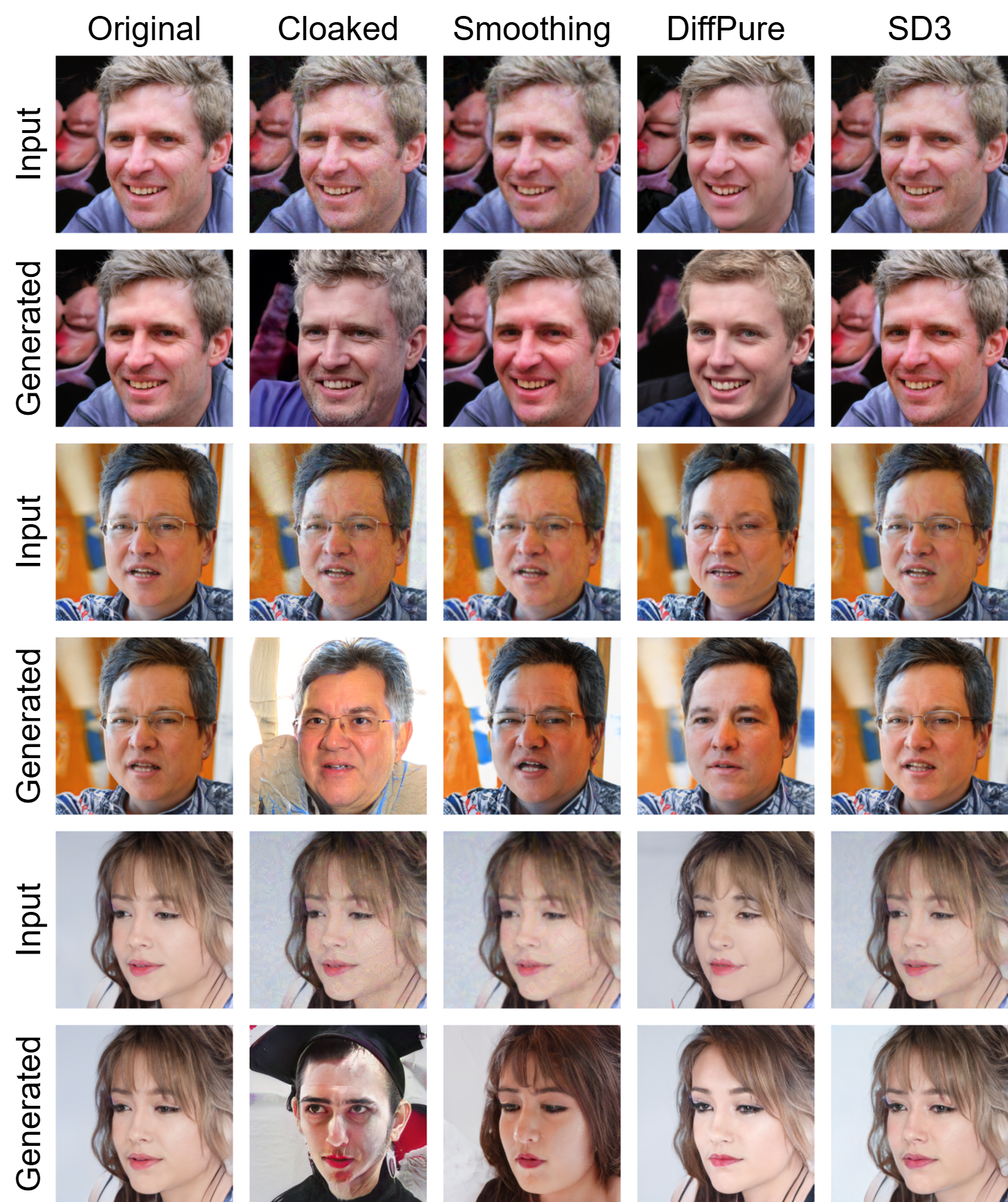} 
\caption{More image samples for UnGANable.}
\label{fig:unganable-images-more}

\end{figure}

\para{Experimental setup.} 

\textit{Defense setup:} The GAN inversion process is performed using StyleGANv2~\cite{karras2020analyzing}. StyleGANv2 is the second iteration of StyleGAN~\cite{karras2019style}, a state-of-the-art model that synthetically generates images using style vectors representing high-level attributes and stochastic variation provided through noise. We choose StyleGANv2 because it is shown to have the best results of UnGANable's tested inversion models.  StyleGANv2 is trained on the Flickr Faces High Quality (FFHQ) dataset~\cite{karras2019style}, a high-quality dataset of various human faces.

When selecting a cloak, we target the black-box setting, where the defender is unaware of the attacker's GAN model or inversion technique.  From the two cloaks that satisfy this setting, we choose Cloak v1 over Cloak v4. The difference between these methods is that Cloak v1 handles optimization-based inversion, where the inversion process uses random noise as a starting point, instead of hybrid inversion, where an encoded variant of the input image is used instead.  The former is the weaker inversion strategy and thus more compelling for showing the attack improvement of our pipeline.

\textit{Attack setup:} We test strength values from 0.025 in increasing increments of 0.025 until we find the optimal strength value. Since 0.05 strength yielded worse utility and performance than 0.025 strength, we choose 0.025. We find that even the most subtle denoising can significantly increase performance. Because UnGANable uses $256 \times 256$ images, an additional Stable Diffusion Upscaler~\cite{rombach2022high} step is integrated into the pipeline to keep consistent with other case studies where our pipeline is used for $512 \times 512$ images. We use Stable Diffusion x4 Upscaler~\cite{rombach2022high} to perform an upscale to $1024 \times 1024$ with the same positive prompt as the denoising model.  We downscale this image to $512 \times 512$ as the input to our pipeline and subsequently downscale the output to $256 \times 256$.

\textit{Baselines for comparison:} We include Gaussian Smoothing and DDPM-based DiffPure~\cite{nie2022diffusion} as baselines. Gaussian Smoothing is a pixel averaging method that is tested in UnGANable as an easy-to-apply countermeasure. The process involves using a filter width of $k$, representing the nearest $k$ pixels to average for each pixel.  We test filter width values in the set \{1,3,5,7,9,11\} following UnGANable's evaluation and report the best results at filter width 3. 

DiffPure is an adversarial purifier that integrates Diffusion models and strategies such as Denoising Diffusion Probabilistic Models (DDPM)~\cite{ho2020denoising}, a pixel space strategy capable of generating high quality images from noise. DDPM-based DiffPure has been trained on CelebA-HQ~\cite{karras2017progressive}, a high-quality dataset of celebrity faces. We run DiffPure for \{200, 300, 400, 500\} timesteps of which we report the best DiffPure results at 200 timesteps.

\textit{Attack evaluation metrics:} The Matching Rate is determined by the number of successfully reconstructed images divided by the number of total images. From UnGANable's implementation, an image is considered successfully reconstructed if the contained faces of the ground truth and reconstructed images have a FaceNet~\cite{schroff2015facenet} similarity distance of 0.58 or less, indicating preserved identity. Based on UnGANable's utility thresholds, an image is considered high-quality if MSE $<$ 0.001 and SSIM $>$ 0.9. FLUX, SD3, and SDXL denoising all successfully meet these thresholds. We measure the utility across all 500 images per setting.

\begin{figure}[t]  
\centering 
\includegraphics[width=8cm]{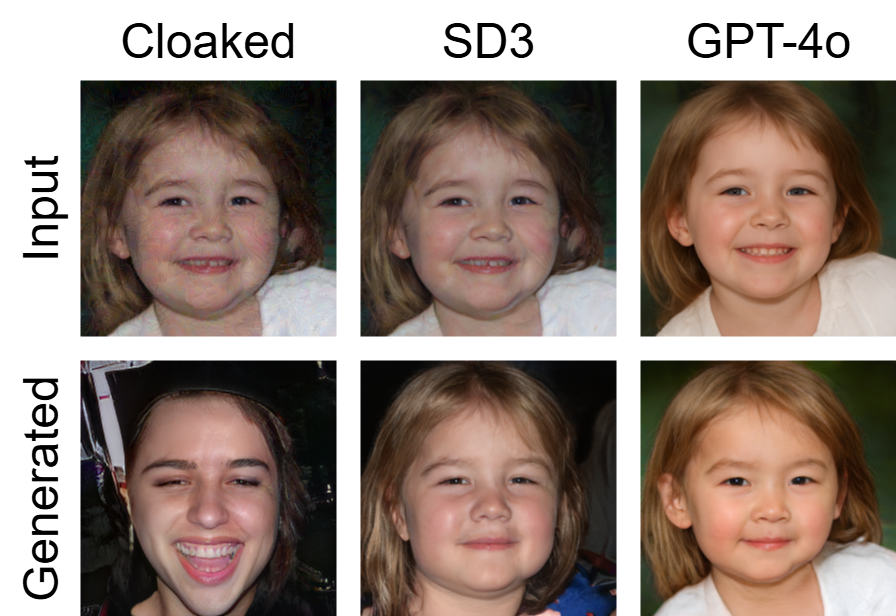} 
\caption{A GPT-4o image sample for UnGANable.  While GPT-4o does not see much improvement in performance, certain face features are recovered during generation better than SD3.}
\label{fig:unganable-images-gpt4o}
\end{figure}

\para{Results.}
\noindent\textbf{\textit{RQ3:}} With GPT-4o on UnGANable, we require the addition of referenceless metrics to better capture the utility of GPT-4o denoised images. We consider the widely-used referenceless BRISQUE~\cite{mittal2012no} but given Case Study 1's focus on face images, a referenceless metric centered on that focus is more appropriate.  SER-FIQ~\cite{terhorst2020ser} is a suitable alternative that outperforms BRISQUE as a referenceless metric for face images.  SER-FIQ uses stochastic embedding robustness to measure the quality of face images. SER-FIQ values are bounded between 0 and 1 with higher values indicating better face image quality.

\subsection{Case Study 2: In-processing Watermark (ICLR’25)}
\label{appendix:attack-study2-settings}

\begin{table}[!t]
\centering
\small
\setlength{\tabcolsep}{1.5pt}
\setlength\extrarowheight{2pt}
\caption{Full denoising results for PRC Watermark on the best prompt (C8). This table includes Regen-VAE C.}
\begin{tabular}{l"c|c|c|c}
\textbf{Attacker}         & \textbf{TPR@FPR ↓} & \textbf{PSNR ↑} & \textbf{SSIM ↑} & \textbf{KID ↓}  \\ \Xhline{1.1pt}
\textbf{No attack}     & 1.000              & -               & -               & 0.0183          \\
\textbf{Smoothing (Avg.)} & 1.000              & \textbf{30.656}          & \textbf{0.868}           & 0.0294          \\
\textbf{DiffPure}         & 0.280              & 21.694          & 0.305           & 0.0302          \\
\textbf{Regen-VAE B}      & 0.312              & 28.859          & 0.770           & 0.0618          \\
\textbf{Regen-VAE C}      & 0.360              & 29.741          & 0.786           & 0.0543          \\ \Xhline{1.1pt}
\textbf{SD1.5}            & 0.878              & 23.314          & 0.611           & 0.0248          \\
\textbf{SDXL}             & \textbf{0.000}     & 24.018          & 0.652           & \textbf{0.0142} \\
\textbf{SD3}              & 0.262              & 25.613 & 0.700  & 0.0196 \\
\rowcolor[gray]{0.85} \textbf{FLUX}             & 0.258     & 28.042 & 0.775  & 0.0196
\end{tabular} 
\label{tab:prc-watermark-full}

\end{table}

\begin{figure}[t]  
\centering 
\includegraphics[width=8.75cm]{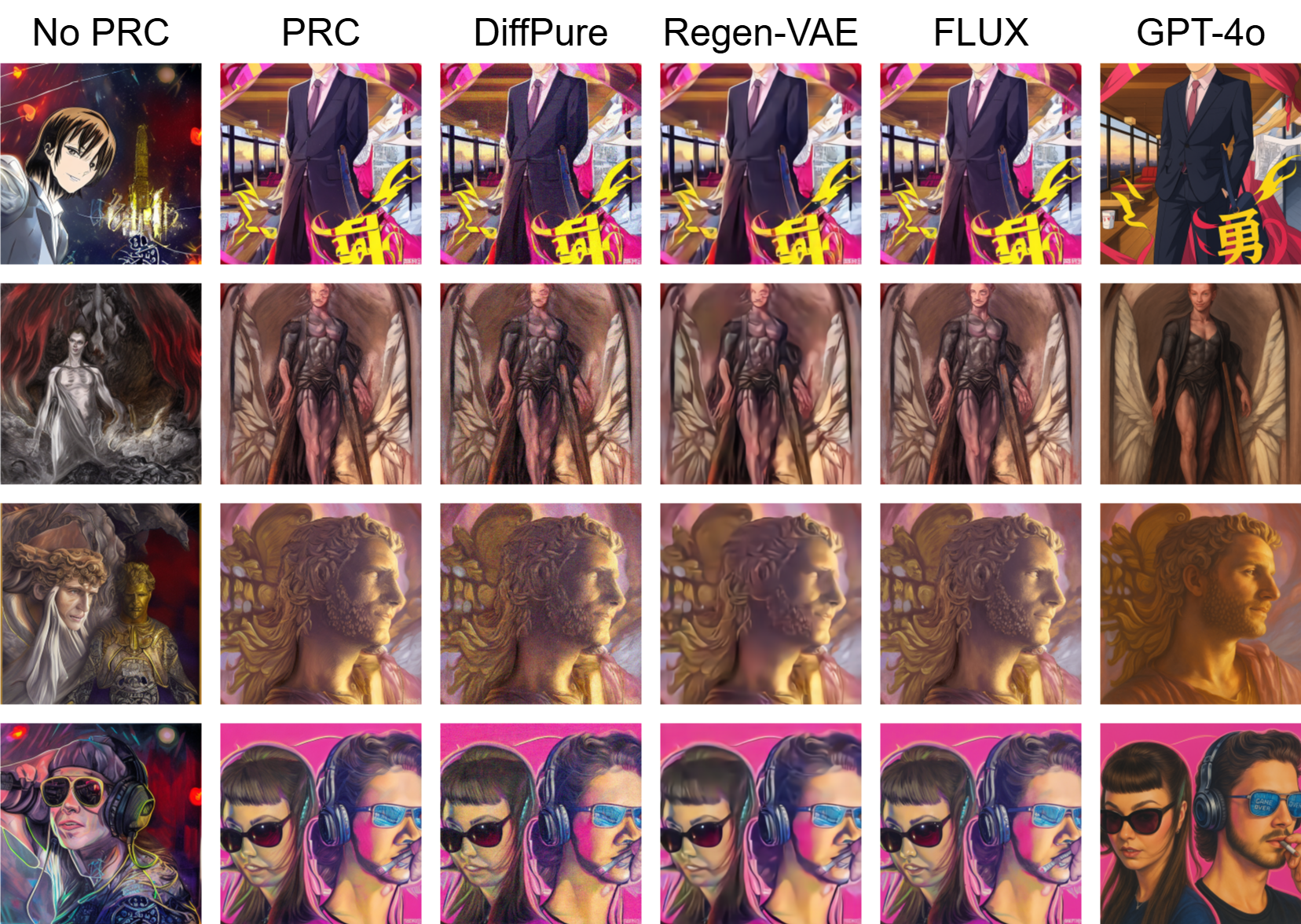} 
\caption{More image samples for PRC Watermark.}
\label{fig:prc-images-more}

\end{figure}

\para{Experimental setup.} 

\textit{Defense setup:} PRC Watermark provides two watermark verification schemes to evaluate performance: \textit{Detect}, which detects the presence of the watermark, and \textit{Decode}, which fully decodes the retrieved watermark. We choose \textit{Detect} as it is more robust to watermark removal attacks. In terms of hyperparameters, the default settings from the provided public implementation are used.  This includes a setting for $t$ which represents the sparsity of the parity checks that PRC Watermark solves by sampling a pseudorandom pattern. From the implementation, we set $t$ to 3, the number of inversion steps to 50 and the False Positive Rate (FPR) value to 0.00001. PRC Watermark is implemented such that the resulting FPR will never exceed the provided rate during generation.  Given our number of images and the default FPR value, this ensures that every generated image is watermarked.

\textit{Attack setup:} Similar to Case Study 1, we test strength values from 0.025 in increasing increments of 0.025 until we find the optimal strength value. With this strategy, we choose a strength of 0.15 for our denoising pipeline for PRC Watermark. This strength setting is sufficient to remove PRC Watermark while maintaining high 
utility.

\textit{Baselines for comparison:}  We include Gaussian Smoothing, DiffPure~\cite{nie2022diffusion}, and Regen-VAE~\cite{zhao2025invisible} as baselines. We perform Gaussian Smoothing with filter widths \{1, 3, 5, 7, 9, 11\}. Unlike Case Study 1, we find that Gaussian Smoothing does not show any improvement for any setting and as such we average the results together. 

For DiffPure, we once again use the provided DDPM implementation.  However, as PRC Watermark does not exclusively use face images, we substitute the CelebA checkpoint for a $512\times512$ ImageNet~\cite{deng2009imagenet} Diffusion checkpoint with the class conditioning steps removed. Removing the class conditioning is necessary as we do not have class labels for any of our Section~\ref{sec:attack-vs-existing-defenses} case studies. We run this implementation with timesteps \{100, 200, 300\} from experimentation and report the best results at 100 timesteps.

Regen-VAE is a regeneration attack against watermarking that is instantiated with a variational autoencoder (VAE). It utilizes pretrained image compression schemes from the Compress AI library~\cite{begaint2020compressai} as VAEs. We report two main variants of Regen-VAE, \textit{Regen-VAE B} and \textit{Regen-VAE C}, based on the following image compression models: the hyperprior variant of Bmshj2018~\cite{balle2018variational} and the anchor variant of Cheng2020~\cite{cheng2020learned}, respectively. These models are chosen because they are the two best state-of-the-art image compression models. Cheng2020 is the better of the two compression schemes as it leverages Gaussian mixture models and attention to improve performance. For both Regen-VAE models, compression factors from 1 through 6 are tested and the best denoising results, corresponding to quality 1 for both variants, are reported.  

\begin{table}[t]
\centering
\small
\setlength{\tabcolsep}{1.5pt}
\setlength\extrarowheight{2pt}
\caption{Denoising results with GPT-4o improvement for PRC Watermark. The 100 remaining images were chosen as the most challenging to denoise.}
\begin{tabular}{l"c|c"c|c}
& \multicolumn{2}{c"}{\textbf{Full Set}} &  \multicolumn{2}{c}{\textbf{100 Remaining Images}} \\ \cline{2-5}
       \textbf{Attacker} & \textbf{\begin{tabular}[c]{@{}c@{}}TPR@\\FPR ↓\end{tabular}} & \textbf{KID ↓} &  \textbf{\begin{tabular}[c]{@{}c@{}}TPR@\\FPR ↓\end{tabular}} & \textbf{BRISQUE ↓}\\ \Xhline{1.1pt}
\textbf{FLUX}      &  0.258 & 0.0196 & 1.000 & 5.164                                                               \\
\rowcolor[gray]{0.85} \textbf{+ GPT-4o} & 0.060 & 0.0174 & 0.010 & 2.859                                   
\end{tabular}

\label{tab:prc-watermark-gpt}

\end{table}

\para{Results.}
\noindent\textbf{\textit{RQ3:}} 
Similarly to Case Study 1, we require referenceless metrics to capture the utility of GPT-4o denoised images. We choose the 100 most easily watermarked images remaining after the attack, after ranking them based on distance to a calculated detection threshold value.  A greater distance indicates greater confidence that the PRC Watermark is present. This is done due to budget restrictions and to show whether or not GPT-4o can break the most confident protections.
Because PRC Watermark's image samples are much more abstract and generalized compared to UnGANable, we use BRISQUE~\cite{mittal2012no} instead of SER-FIQ.  BRISQUE is a referenceless image quality evaluator that scores with a bounded value from 0 to 100 where a lower number indicates a higher quality image. BRISQUE has been used as an evaluation metric in other significant related work~\cite{van2023anti}. We report the BRISQUE score for a set of images as the average score of that set.

\subsection{Case Study 3: Post-processing Watermark (ICLR'25)}\label{appendix:attack-study3-settings}
\begin{table}[!t]
\centering
\small
\setlength{\tabcolsep}{1.5pt}
\setlength\extrarowheight{2pt}
\caption{Full denoising results for VINE on the best prompt (C6).  The performance of SD1.5, SDXL, SD3, Regen-VAE C, and Gaussian Smoothing are included.}
\begin{tabular}{l"c|c|c|c|c}
\textbf{Attacker}                & \textbf{\begin{tabular}[c]{@{}c@{}}TPR@\\FPR ↓\end{tabular}} & \textbf{PSNR ↑} & \textbf{SSIM ↑} & \textbf{LPIPS ↓} & \textbf{KID ↓}  \\ \Xhline{1.1pt}
\textbf{No Denoising}            & 1.000              & 37.479          & 0.993           & 0.007            & 0.0000          \\
\textbf{Smoothing (Avg.)}        & 1.000              & \textbf{32.524}          & \textbf{0.917}           & 0.182            & 0.0015          \\
\textbf{DiffPure}                & 0.991              & 22.087          & 0.336           & 0.575            & 0.0033          \\
\textbf{Regen-VAE B}             & 0.976              & 29.424          & 0.832           & 0.275            & 0.0194          \\
\textbf{Regen-VAE C}             & 0.982              & 29.949          & 0.840           & 0.240            & 0.0182          \\
\textbf{Sto. Regen.} & 0.981              & 22.325          & 0.636           & 0.270            & 0.0106          \\
\textbf{Det. Inver.} & 0.997              & 25.408          & 0.750           & 0.225            & 0.0072          \\ \Xhline{1.1pt}
\textbf{SD1.5}                   & 0.986              & 22.786          & 0.649           & 0.208            & \textbf{0.0005} \\
\textbf{SDXL}                    & \textbf{0.720}     & 22.998          & 0.675           & 0.227            & 0.0037          \\
\textbf{SD3}                     & 0.817     & 23.867 & 0.690           & 0.185   & \textbf{0.0005} \\
\rowcolor[gray]{0.85} \textbf{FLUX}                    & 0.878              & 26.646 & 0.775  & \textbf{0.154}   & 0.0006
\end{tabular}
\label{tab:vine}
\end{table}

\begin{figure}[t]
\centering
\includegraphics[width=8.75cm]{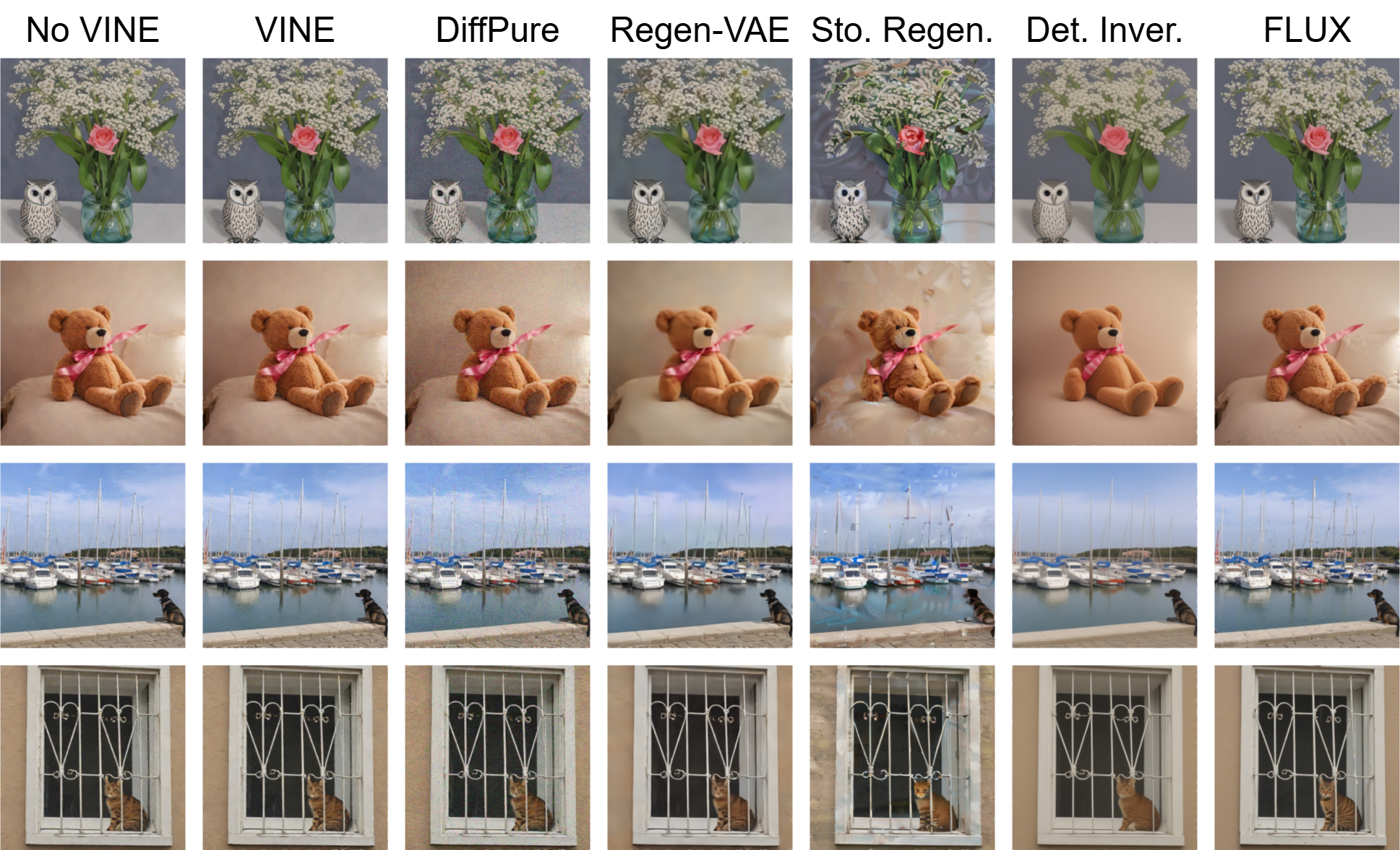}
\caption{Image samples for VINE.  FLUX is able to improve performance without sacrificing significant utility.}
\label{fig:vine-images}
\end{figure} 

\para{Experimental setup.} 

\textit{Defense setup:} To apply the watermark, VINE has two variants: the base model, VINE-B, and a fine-tuned, more robust model, VINE-R.  We use the more robust variant as used in the original work. We also use their provided pretrained checkpoints for the VINE-R encoder and decoder. The encoded message used in the watermark is set to the default ``Hello World!''.

\textit{Attack setup:} From experimentation, our denoising pipeline is tested starting with strength 0.05 in increments of 0.05. We report the results of strength 0.25, which enables our pipeline to completely outperform all baselines while still providing competitive utility.

\textit{Baselines for comparison:} Stochastic Regeneration adds noise to an image and then denoises it using a Diffusion model given a number of timesteps. It functions similarly to DiffPure with SD2.1 as the Diffusion method. Deterministic Inversion inverts a clean image into a noisy image which is then used to reconstruct the clean image given a number of sampling steps.  SD2.1 is also used as the base model for Deterministic Inversion. 

Following VINE's evaluation, we perform the Stochastic Regeneration baseline from timesteps 60 to 240 in increments of 20 of which we report the best setting.  We similarly report the best setting for Deterministic Inversion, which is tested from sampling steps 15 to 45 in increments of 10.  We determine the best settings to be 240 timesteps for Stochastic Regeneration and 15 sampling steps for Deterministic Inversion. We implement Regen-VAE, DiffPure, and Gaussian Smoothing similarly to Case Study 2, with the only exception that the filter width of 11 is not tested for Gaussian Smoothing as VINE also did not test it.

\begin{table}[!t]
\centering
\small
\setlength{\tabcolsep}{1.5pt}
\setlength\extrarowheight{2pt}
\caption{Full denoising results for SIREN on the best prompt (C6).  The performance of Regen-VAE B is included.}
\begin{tabular}{l"c|c|c|c|c}
                      \textbf{Attacker} & \textbf{TPR@Sign. ↓} & \textbf{PSNR ↑} & \textbf{SSIM ↑} & \textbf{KID ↓} & \textbf{LPIPS ↓} \\ \Xhline{1.1pt}
\textbf{No Denoising} & 1.000              & 39.142          & 0.880           & 0.070          & 0.016            \\
\textbf{DiffPure}     & 0.101              & 29.835          & 0.596           & 0.084          & 0.215            \\
\textbf{Regen-VAE B}  & 0.652              & 30.552          & 0.702           & \textbf{0.062}          & 0.110            \\
\textbf{Regen-VAE C}  & 0.591             & \textbf{31.886}          & \textbf{0.890}           & 0.071         & 0.085            \\ \Xhline{1.1pt}
\textbf{SD1.5}        & 0.147              & 22.348          & 0.803  & 0.100 & 0.122            \\
\textbf{SDXL}         & \textbf{0.000}     & 22.049         & 0.765           & 0.100          & 0.136            \\
\textbf{SD3}          & 0.001              & 22.935 & 0.612          & 0.112          & 0.121   \\
\rowcolor[gray]{0.85} \textbf{FLUX}         & 0.016     & 28.882 & 0.787  & 0.071 & \textbf{0.050}  
\end{tabular}
\label{tab:siren-full}
\end{table}

\begin{figure}[t]
\centering
\includegraphics[width=8.75cm]{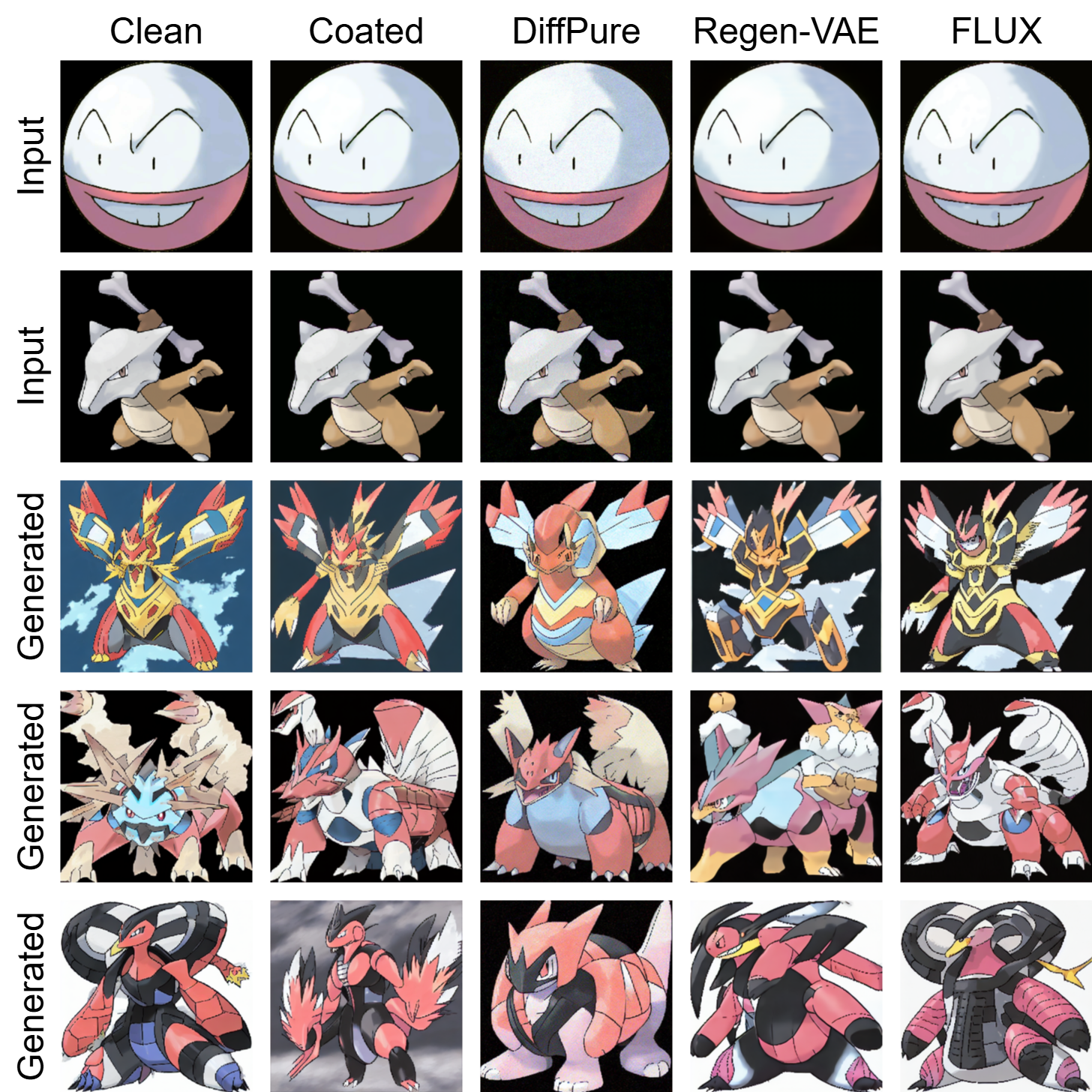}
\caption{Image samples for SIREN.}
\label{fig:siren-images}

\end{figure}

\subsection{Case Study 4: Verifying Unauthorized Data Usage in Personalized Models (IEEE S\&P'25)}
\label{appendix:attack-study4-settings}

\para{Experimental setup.} 

\textit{Defense setup:} For the SIREN encoder (coating mechanism) and decoder (verification scheme), we use pre-trained checkpoints (fine-tuned for the Pokemon dataset) provided by the authors, configured with the recommended hyperparameters.  We use SIREN's provided Pokemon encoder checkpoint to encode the traceability cloak and the Pokemon decoder to detect the coating within generated images.  We use SIREN's provided script to generate fine-tuned LoRA~\cite{hu2022lora} checkpoints with pretrained SD1.5 as a starting point.  This is done for 80 epochs and with the learning rate set to 0.0001.  We use the default parameters when using the fine-tuned checkpoint to generate 1000 images for each denoising setting.

\textit{Attack setup:} Our denoising pipeline is tested from strengths 0.05 to 0.45 in increments of 0.1. We report the results for strength 0.35 from experimentation.  This strength provided performance values close to 0 TPR@Significance while maintaining high utility.

\textit{Attack evaluation metrics:} Following SIREN, we use a sample size of 30 for the generated images and perform the Kolmogorov-Smirnov test 10,000 times to obtain TPR@Significance.

\subsection{Case Study 5: INSIGHT (USENIX Security'24)}
\label{appendix:attack-study5-settings}

\begin{figure}[t]
\centering
\includegraphics[width=8.75cm]{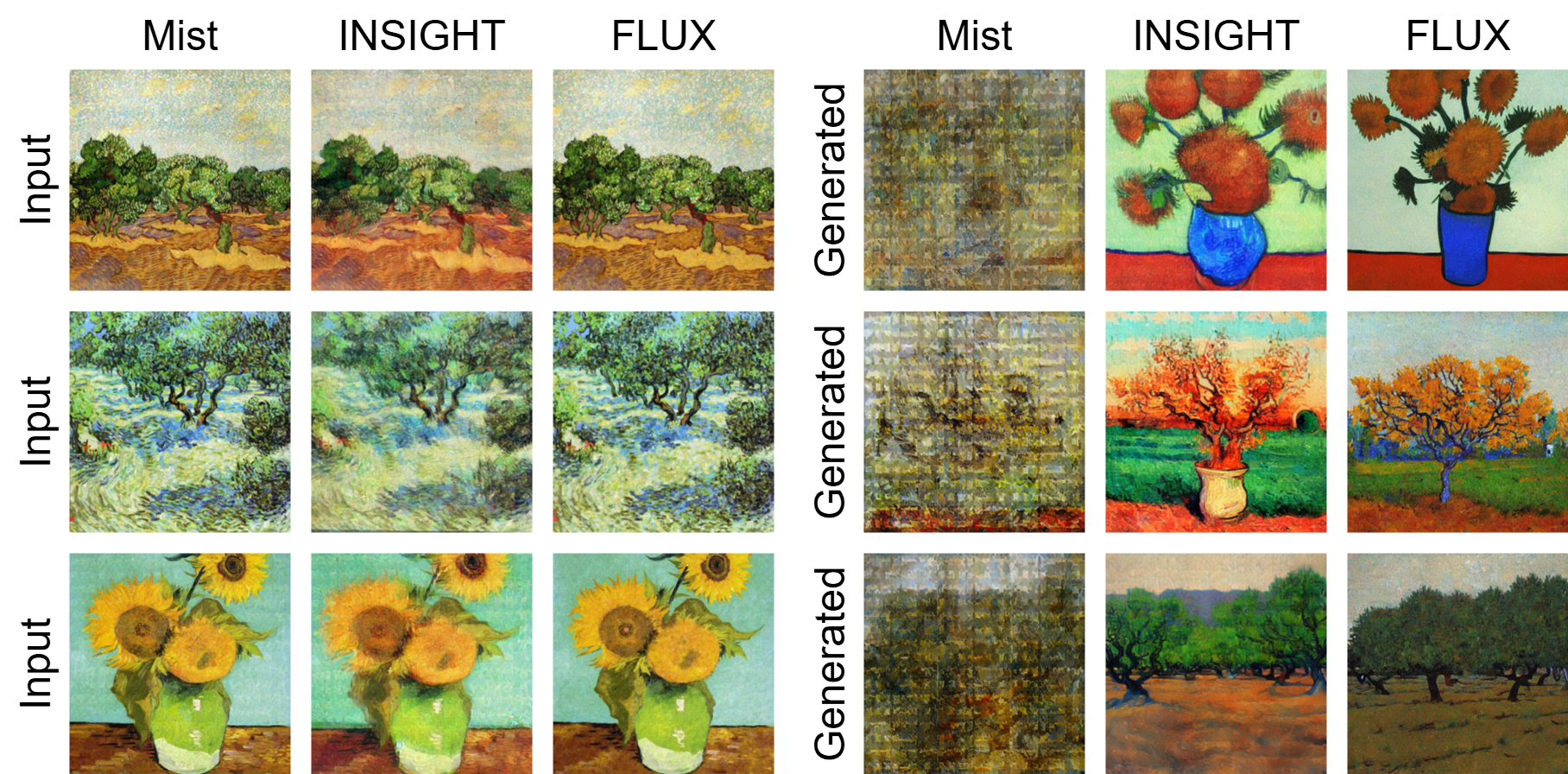} 
\caption{More samples for INSIGHT.}
\label{fig:insight-images-more}

\end{figure}

\para{Experimental setup.}

\textit{Attack setup.} Following their codebase configuration, we run INSIGHT for 500 iterations and follow their selection of loss weights.  The total loss consists of four components: the visual bound, the UNet alignment loss, the VAE alignment loss, and the reconstruction loss. From their implementation, the weights for these losses are set to 0.1, 0.01, 1, and 1, respectively.

We use the best FLUX setting from PRC Watermark, a latent space protection scheme much like Mist, consisting of prompt C8 and a strength of 0.15. We use these denoised images to fine-tune an SD1.5 model with DreamBooth~\cite{ruiz2023dreambooth} training and subsequently generate 1000 images. We keep the settings of DreamBooth consistent for all attacks using a learning rate of $5e^{-6}$ and maximum train steps of 400 (from Hugging Face).

\textit{Evaluation metrics:} We perform CLIP accuracy using the top 40 most common WikiArt~\cite{artgan2018} styles as classes, with Van Gogh's art style being the desired class ``post impressionism''. We use the standard ViT-B/32 model as our pretrained CLIP model to perform this classification.

\subsection{Case Studies 6 and 7: Noisy Upscaling (ICLR'25) and LightShed (USENIX Security'25)}
\label{appendix:attack-study6-and-7-settings}

\para{Experimental setup.}

\noindent\textit{Attack setup.} We use Mist's latest implementation, Mist v3, as our protection scheme. We use the default parameters for Mist generation including a strength of 16, 100 optimization steps, the ``fused'' mode for watermarking, and a fused weight of 1. We also set the image size to be consistently $256 \times 256$ throughout this case study following Mist's text-to-image generation experiment.

Mist's evaluation utilizes the LSUN-cat~\cite{yu2015lsun} dataset but it is no longer publicly available.  As a result, we adopt the higher quality LAION-Aesthetic dataset~\cite{schuhmann2022laion,laion_ev_2025} and filter 100 cat images. Specifically, we filter unique images that contain ``cat'' in their associated captions and have good CLIP scores (using a score threshold of 0.24) corresponding to ``a photo of a cat.'' 

For finding the pseudo-word, we use Textual Inversion~\cite{gal2022image}'s provided pretrained checkpoints as a starting point to fine-tune on. We then use the resulting fine-tuned checkpoint as the model to generate images adhering to the pseudo-word. For generation, we use the default parameters from Textual Inversion including a guidance scale of 10 and 50 DDIM steps.

For LightShed, we assume that every input image is poisoned and change the output image size from $512\times512$ to $256\times256$. We use their Autoencoder checkpoint trained on NightShade~\cite{shan2024nightshade}, Glaze~\cite{shan2023glaze}, Mist, and MetaCloak~\cite{liu2024metacloak}. We use the default hyperparameters for LightShed.

For Noisy Upscaling, we use LightShed's implementation of Noisy Upscaling including its default hyperparameters of a 0.05 standard deviation for the Noise step and a noise level of 160 for the Upscale step. We additionally add a non-adversarial setting only consisting of Mist protection as a baseline for further insights.

Using FLUX with prompt C8 and strength 0.15 from PRC Watermark as a starting point, we adjust the strength and inference steps hyperparameters until we obtain FLUX with a strength of 0.35 and inference steps set to 100. Our reasoning for adjusting the timesteps was to determine whether we can improve the quality of generated images after the denoise.  We choose the number of inference steps from \{28, 50, 100, 300, 500\} with 28 being the default number of timesteps for FLUX.

Since we use $256 \times 256$ for this case study, we apply an upscaling step prior to our denoising (both GPT-4o and FLUX) to scale the input image to $512 \times 512$. We follow the same Stable Diffusion Upscaler~\cite{rombach2022high} and scaling strategy as in Case Study 1 (Appendix~\ref{appendix:unganable-settings}).

\subsection{Details of User Study for Case Study 6 and 7.}\label{appendix:userstudy}

\begin{table*}[t]
\small
\centering
\setlength{\tabcolsep}{1.5pt}
\setlength\extrarowheight{2pt}
\caption{Proportion of image pairs in Study 1 where the clean image has better utility than the corresponding counterpart generated image given a Mist-protected or denoised input. ${}^\textbf{**}$ indicates that the corresponding proportion is statistically significant with $p < 0.001$ and ${}^\textbf{***}$ indicates $p < 0.0001$.}
\begin{tabular}{ll"llllll}
 &
   &
  \multicolumn{6}{c}{\textbf{Proportion of image pairs where clean image has better utility}} \\ \cline{3-8} 
 &
   &
  \multicolumn{2}{c"}{\textbf{Quality metrics ↑}} &
  \multicolumn{4}{c}{\textbf{Concept-appropriateness metrics ↑}} \\ \cline{3-8} 
 &
   &
  \multicolumn{1}{c|}{\textbf{Noise}} &
  \multicolumn{1}{c"}{\textbf{Artifacts}} &
  \multicolumn{1}{c|}{\textbf{Detail}} &
  \multicolumn{1}{c|}{\textbf{Concept fit}} &
  \multicolumn{1}{c|}{\textbf{Prompt fit}} &
  \multicolumn{1}{c}{\textbf{Overall realism}} \\ \Xhline{1.1pt} 
\rowcolor[gray]{0.85} \multicolumn{1}{l}{} &
  \textbf{Clean image vs. Mist + GPT-4o} &
  \multicolumn{1}{l|}{0.08${}^\textbf{***}$} &
  \multicolumn{1}{l"}{0.29} &
  \multicolumn{1}{l|}{0.10${}^\textbf{***}$} &
  \multicolumn{1}{l|}{0.00${}^\textbf{***}$} &
  \multicolumn{1}{l|}{0.10${}^\textbf{***}$} &
  0.18${}^\textbf{**}$ \\
\multicolumn{1}{l}{} &
  \textbf{Clean image vs. Mist + FLUX} &
  \multicolumn{1}{l|}{0.95${}^\textbf{***}$} &
  \multicolumn{1}{l"}{0.33} &
  \multicolumn{1}{l|}{0.20} &
  \multicolumn{1}{l|}{0.42} &
  \multicolumn{1}{l|}{0.36} &
  0.30 \\
\multicolumn{1}{l}{} &
  \textbf{Clean image vs. Mist + LightShed} &
  \multicolumn{1}{l|}{0.94${}^\textbf{***}$} &
  \multicolumn{1}{l"}{0.88${}^\textbf{***}$} &
  \multicolumn{1}{l|}{0.85${}^\textbf{***}$} &
  \multicolumn{1}{l|}{0.90${}^\textbf{***}$} &
  \multicolumn{1}{l|}{0.90${}^\textbf{***}$} &
  0.90${}^\textbf{***}$ \\ 
\multicolumn{1}{l}{} &
  \textbf{Clean image vs. Mist + Noisy Upscaling} &
  \multicolumn{1}{l|}{1.00${}^\textbf{***}$} &
  \multicolumn{1}{l"}{0.92${}^\textbf{***}$} &
  \multicolumn{1}{l|}{0.93${}^\textbf{***}$} &
  \multicolumn{1}{l|}{0.60} &
  \multicolumn{1}{l|}{0.58} &
  0.85${}^\textbf{***}$ \\
\multicolumn{1}{l}{} &
  \textbf{Clean image vs. Mist} &
  \multicolumn{1}{l|}{1.00${}^\textbf{***}$} &
  \multicolumn{1}{l"}{0.40} &
  \multicolumn{1}{l|}{0.59} &
  \multicolumn{1}{l|}{0.38} &
  \multicolumn{1}{l|}{0.31} &
  0.44
\end{tabular}
\label{tab:study1-input-utility-comparison}
\end{table*}

\begin{table*}[t]
\small
\centering
\setlength{\tabcolsep}{1.5pt}
\setlength\extrarowheight{2pt}
\caption{Proportion of image pairs in Study 2 where the generated image when attacked with GPT-4o has better utility than the corresponding image attacked with Noisy Upscaling or LightShed. ${}^\textbf{***}$ indicates that the corresponding proportion is statistically significant with  $p < 0.0001$.}
\begin{tabular}{ll"llllll}
 &
   &
  \multicolumn{6}{c}{\textbf{Proportion of image pairs where GPT-4o image has better utility}} \\ \cline{3-8} 
 &
   &
  \multicolumn{2}{c"}{\textbf{Quality metrics ↑}} &
  \multicolumn{4}{c}{\textbf{Concept-appropriateness metrics ↑}} \\ \cline{3-8} 
 &
   &
  \multicolumn{1}{c|}{\textbf{Noise}} &
  \multicolumn{1}{c"}{\textbf{Artifacts}} &
  \multicolumn{1}{c|}{\textbf{Detail}} &
  \multicolumn{1}{c|}{\textbf{Concept fit}} &
  \multicolumn{1}{c|}{\textbf{Prompt fit}} &
  \multicolumn{1}{c}{\textbf{Overall realism}} \\ \Xhline{1.1pt} 
\multicolumn{1}{l}{} &
  \textbf{GPT-4o vs. LightShed} &
  \multicolumn{1}{l|}{1.00${}^\textbf{***}$} &
  \multicolumn{1}{l"}{0.94${}^\textbf{***}$} &
  \multicolumn{1}{l|}{0.87${}^\textbf{***}$} &
  \multicolumn{1}{l|}{1.00${}^\textbf{***}$} &
  \multicolumn{1}{l|}{0.94${}^\textbf{***}$} &
  0.94${}^\textbf{***}$ \\  
\multicolumn{1}{l}{} &
  \textbf{GPT-4o vs. Noisy Upscaling} &
  \multicolumn{1}{l|}{1.00${}^\textbf{***}$} &
  \multicolumn{1}{l"}{0.89${}^\textbf{***}$} &
  \multicolumn{1}{l|}{1.00${}^\textbf{***}$} &
  \multicolumn{1}{l|}{1.00${}^\textbf{***}$} &
  \multicolumn{1}{l|}{1.00${}^\textbf{***}$} &
  0.95${}^\textbf{***}$  
\end{tabular}
\label{tab:study2-gpt4o-utility-comparison}

\end{table*}

\begin{figure}[t]
\centering
\includegraphics[width=8.75cm]{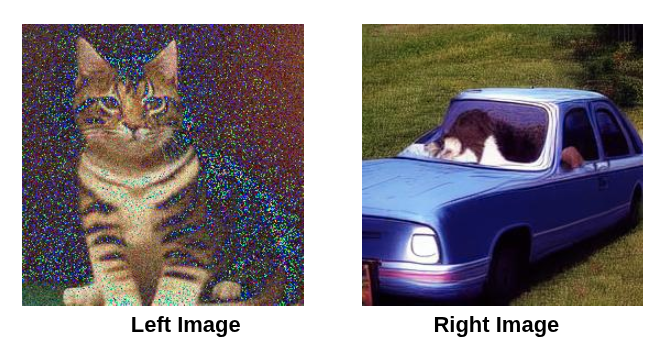} 
\caption{An example of a training phase pair.  The right image is an example of a control image as it is not concept-appropriate, showing a car instead of a cat.}
\label{fig:training-phase-pair}

\end{figure}

\begin{figure}[t]
\centering
\includegraphics[width=8.75cm]{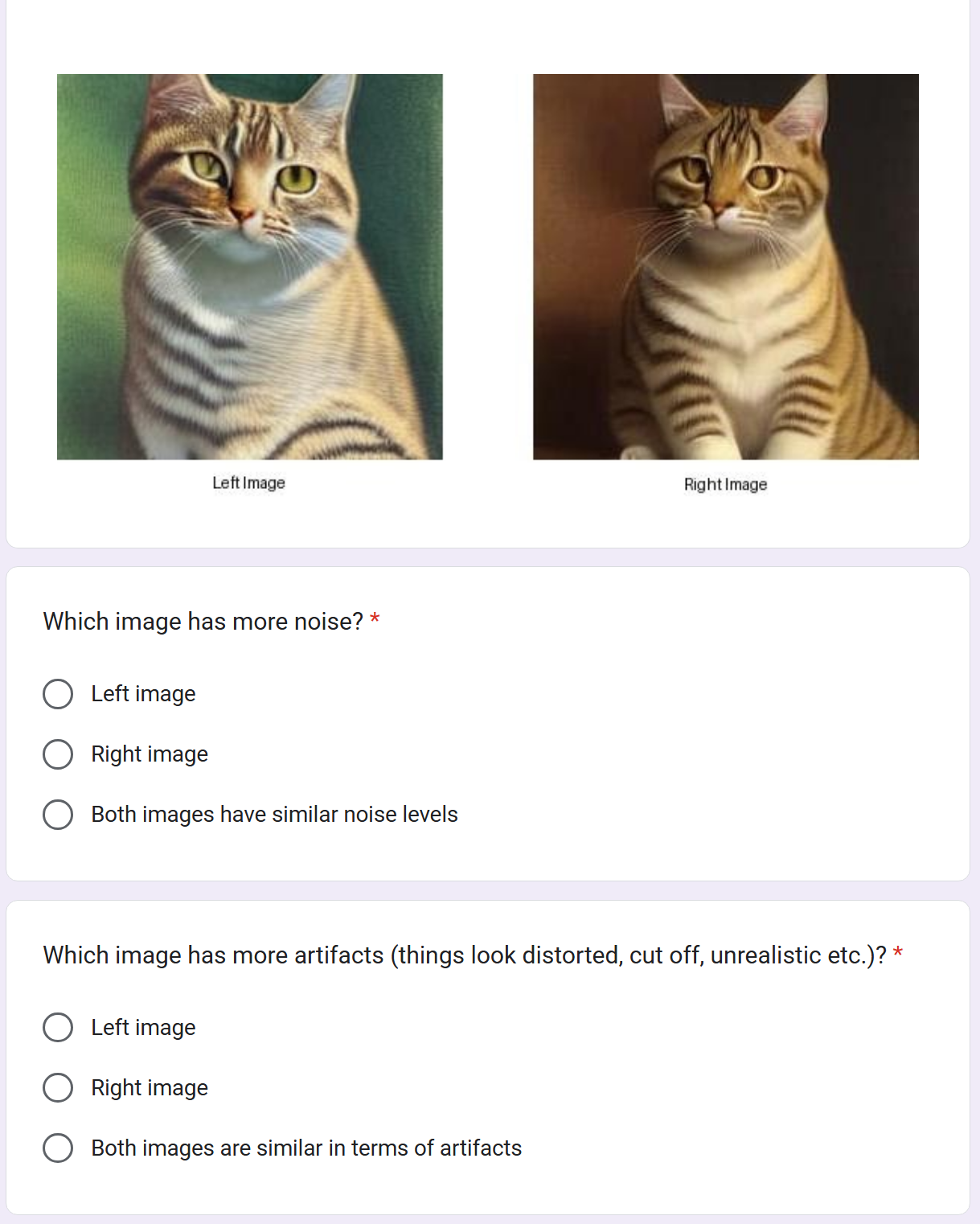} 
\caption{An example image pair and 2 of the 6 questions presented to participants. The right image is the image generated with GPT-4o, which the participant is unaware of.}
\label{fig:survey}

\end{figure}

\para{Study design.} Our studies focused on detecting the better image based on image quality (\textit{noise levels, presence of artifact}) and concept-appropriateness (\textit{details of the image, fit with description of ``cat'', prompt response appropriateness and overall realism}). Our study protocols were approved by the institution IRB of the author(s) who performed the studies.

Each study has two sections: (i) \textbf{a training phase}, where participants were shown three examples (image pairs labeled as ``left'' image and ``right'' image) to explain the image labeling guidelines (Figure~\ref{fig:training-phase-pair} shows an image pair from the training phase), (ii) \textbf{a labeling phase}, where participants were asked to provide their judgment on image quality and concept-appropriateness for multiple image pairs. 

For each image pair, the participant answered the following six questions in line with LightShed~\cite{foerster2025lightshed}:
    \begin{enumerate}
        \item \textbf{Noise:} Which image has more noise?
        \item \textbf{Artifacts:} Which image has more artifacts (things look distorted, cut off, unrealistic etc.)?
        \item \textbf{Detail:} Which image has more details?
        \item \textbf{Concept fit:} Which image fits the description of a “cat” better? 
        \item \textbf{Prompt fit:} Overall, ignoring the image quality, which image is more appropriate as a response to the prompt "photo of a cat" given to AI?
        \item \textbf{Overall realism:} Based on noise, artifacts, detail, prompt response appropriateness, and your impression, which image looks more like a realistic photo of a cat?
    \end{enumerate}

Each of these questions have the following options for the participants:

\begin{itemize}
    \item Left Image
    \item Right Image
    \item Both images [have similar noise levels / are similar in terms of artifacts / have similar level of details / fit the description of a cat equally / are similarly appropriate / are of similar quality]
\end{itemize}

Figure~\ref{fig:survey} shows an example image pair that is presented to the participants for utility evaluation in our survey.

\para{Recruitment.} We recruited 21 participants, 15 for the first study and 6 for the second study. We ensured that participant pool was mutually exclusive across studies through Prolific recruitment filter. We recruited participants whose first language was English and had an approval rating of at least $50\%$ on Prolific~\cite{prolific:online}. The average time to complete the studies was 20.9 minutes and the participants were each paid 5 USD.

\para{Generating data for user studies.} We randomly sample a subset of 110 generated images from the set of 1000 images generated with clean input.  For 100 images of this subset, we sample their counterpart generated images given a Mist-protected or denoised input.  We sample 20 images from Mist protection, 20 images from LightShed, 20 images from Noisy Upscaling, 20 images from FLUX, and 20 images from GPT-4o in this manner.  The remaining 10 images are paired with 10 control images. 

We sample 2 counterpart images from GPT-4o and apply significant Gaussian Noise to serve as control images for Noise.  We sample another 2 counterpart images from GPT-4o and apply significant Gaussian Blurring to serve as control images for Detail.  The remaining 6 images are completely unrelated and chosen from various sources such as INSIGHT~\cite{an2024rethinking} generation, dog images with NightShade~\cite{shan2024nightshade} poisoning, and Textual Inversion~\cite{gal2022image} personalization from MSCOCO~\cite{lin2014microsoft} cat images. These serve as control images for Concept fit, Prompt fit, and Overall realism, with the Overall realism control images also having Gaussian Blur and Gaussian Noise applied.

For the GPT-4o vs. LightShed and GPT-4o vs. Noisy Upscaling experiments, we sample an additional 80 images, 40 from GPT-4o, 20 from LightShed, and 20 from Noisy Upscaling.  They are sampled so that each pair contains two images from denoised inputs that are counterparts of the same clean image.

\para{Ensuring data quality.} We included multiple attention checks and control image pairs to verify that users answered correctly.

\begin{enumerate}

    \item Checkboxes were put at the beginning of \textit{labeling phase} of the study for participants to acknowledge they have understood the instructions in the training phase and are now ready to being the labeling phase.

    \item Random attention check questions places throughout the studies --{\it Which number is even?, Please select "Yes" as the answer to this question.,} and {\it Which century are we living in?}.

    \item Finally, the randomly introduced control image pairs acted as additional quality control. The answer to these pairs was obvious and objective: either the left image or the right image was higher quality, but never both.
\end{enumerate}

The responses of the participants passed all attention checks and were correct for the control images.

\para{Data analysis.} For each image pair, we used majority voting to get the final rating of the image against each evaluation criteria for every image pair. We then calculate the proportion of image pairs where majority voted the clean image as having better quality (for first study) or image generated when attacked with GPT-4o with better quality (for second study). We excluded control image pairs from the analysis. We used one-proportion z-test~\cite{ztest-proportion-test} to check the significance level of this accuracy.

\para{Results.} The full results for the first and second studies are presented in Tables~\ref{tab:study1-input-utility-comparison} and~\ref{tab:study2-gpt4o-utility-comparison}, respectively. 

\begin{table}[t!]
\centering
\small
\setlength{\tabcolsep}{1.5pt}
\setlength\extrarowheight{2pt}
\caption{Full denoising results for UnMarker, FLUX, and GPT-4o on TRW. This table shows both the 0\% and 10\% crop settings.}
\begin{tabular}{l"cc"ccc}
                        & \multicolumn{2}{c"}{\textbf{Performance}}                                                                     & \multicolumn{2}{c}{\textbf{\begin{tabular}[c]{@{}c@{}}Denoising Utility\end{tabular}}}                                              \\ \cline{2-5} 
                        \textbf{Protection} & \multicolumn{1}{c|}{\textbf{\begin{tabular}[c]{@{}c@{}}Inv.\\ Dist. ↓\end{tabular}}} & \textbf{\begin{tabular}[c]{@{}c@{}}TPR@MAE\\ = 68.48 ↓\end{tabular}} 
                        & \multicolumn{1}{c|}{\textbf{\begin{tabular}[c]{@{}c@{}}CLIP\\ FID ↓\end{tabular}}} 
                        & \textbf{BRISQUE ↓} \\ \Xhline{1.1pt}
\textbf{UnMarker (HL)}  & \multicolumn{1}{c|}{0.0162}              & 0.9011                                                             
& \multicolumn{1}{c|}{6.774}             
& 62.969 \\
\textbf{UnMarker (L)}   & \multicolumn{1}{c|}{0.0166}              & 0.9560                                                             
& \multicolumn{1}{c|}{\textbf{5.771}}    
& \textbf{7.728} \\
\textbf{FLUX}           & \multicolumn{1}{c|}{0.0153}              & 0.7912                                                             
& \multicolumn{1}{c|}{11.549}            
& 13.958 \\
\rowcolor[gray]{0.85} \textbf{GPT-4o}         & \multicolumn{1}{c|}{\textbf{0.0149}}     & \textbf{0.6813}                                                    
& \multicolumn{1}{c|}{12.559}            
& 8.002 \\ \Xhline{1.1pt}
\textbf{UnMarker (CHL)} & \multicolumn{1}{c|}{\textbf{0.0140}}     & \textbf{0.1758}                                                    
& \multicolumn{1}{c|}{8.785}             
& 67.552 \\
\textbf{UnMarker (CL)}  & \multicolumn{1}{c|}{0.0142}              & 0.2857                                                             
& \multicolumn{1}{c|}{\textbf{8.106}}    
& 16.070 \\
\textbf{Crop + FLUX}    & \multicolumn{1}{c|}{0.0148}              & 0.6044                                                             
& \multicolumn{1}{c|}{12.429}            
& 23.441 \\
\rowcolor[gray]{0.85} \textbf{Crop + GPT-4o}  & \multicolumn{1}{c|}{0.0146}              & 0.5604                                                             
& \multicolumn{1}{c|}{13.209}            
& \textbf{7.904}
\end{tabular}
\label{tab:unmarker-trw}
\end{table}

\begin{figure}[t]
\centering
\includegraphics[width=8.75cm]{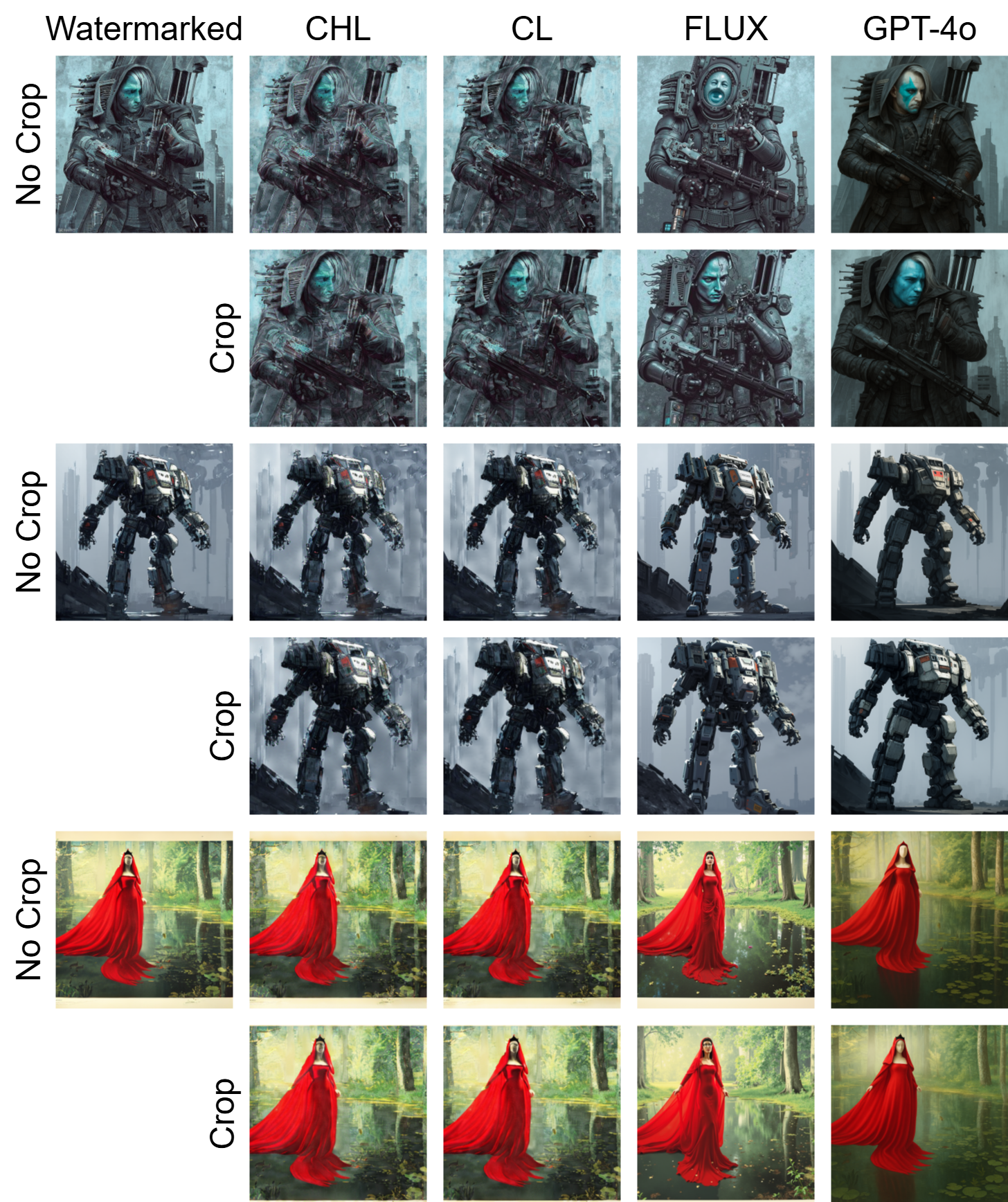}
\caption{More examples for UnMarker.}
\label{fig:unmarker-images-more}
\end{figure}

\subsection{Case Study 8: UnMarker (IEEE S\&P'25)}
\label{appendix:attack-study9-settings}

\para{Background.} Tree-Ring Watermarking (TRW)~\cite{wen2023tree} is a sophisticated watermarking method that embeds a watermarking key into the Fourier Transform of an initial noise vector prior to image generation, making it robust and imperceptible.  Alongside TRW, UnMarker also studies the semantic watermarking method StegaStamp~\cite{tancik2020stegastamp}, a watermarking method that robustly encodes hyperlink bitstrings into images imperceptibly. We choose TRW as the semantic scheme to attack due to StegaStamp's checkpoints being unavailable.

\para{Experimental setup.}

\textit{Attack setup.} 
We follow the implementation of TRW based on UnMarker's codebase which shares the same hyperparameters as TRW's own implementation, including the choice of TRW-Rings, a watermarking radius of 10, a circle for the shape of the mask, and no additional image distortion.  TRW-Rings is the best performing watermark variant of TRW. We follow the provided ``TreeRing.yaml'' file to set UnMarker's hyperparameters.

As per UnMarker's evaluation, we use a dataset of 100 SDP~\cite{stablediffusionprompts:online} prompts to generate watermarked and unwatermarked images.  Of this set, we work with a subset of 91 generated images that can be GPT-4o denoised.  Unfortunately, 9 of the 100 watermarked images, despite all being safe, were falsely blocked by GPT-4o's moderation system as unsafe input.

Using FLUX on C8 with strength 0.15 from Case Study 2, we adjust the strength until we obtain a strength of 0.45.  This strength allows us to significantly improve performance compared to UnMarker in the no crop setting.

\textit{Evaluation metrics.}  UnMarker uses a different threshold than other papers to calculate TPR@FPR (lower is better). Specifically, UnMarker leverages Mean Absolute Error (MAE) to determine whether or not a watermark is considered detected. From experimentation, we select an MAE of 68.48 which is the closest value to the hundredth higher than the lowest MAE of one unwatermarked image (68.47), falsely flagging it as watermarked (i.e. FPR=0.01).

\subsection{Countermeasure Study 1: Denoising-Aware UnGANable}\label{appendix:unganable-countermeasure}

\para{Countermeasure setup.} UnGANable's Cloak v1 is defined by Equation~\ref{eq:cloakv1} where $F$ is the feature extractor and $\mathcal{L}_{total} = \mathcal{L}_{mse}-\mathcal{L}_{cos}$ with $\mathcal{L}_{mse}$ being an MSE similarity loss representing visual similarity and $\mathcal{L}_{cos}$ being a cosine similarity loss representing perceptual or feature similarity. 

\begin{equation}
\text{max}_\textbf{\^{x}}\mathcal{L}_{total}(F(\textbf{\^{x}}),F(\textbf{x}))\text{ s.t. }|\textbf{\^{x}}-\textbf{x}|_\infty < \epsilon 
\label{eq:cloakv1}
\end{equation}

The goal is to find a cloaked image \textbf{\^{x}} such that it is visually similar to \textbf{x} but their feature spaces, $F(\textbf{\^{x}})$ and $F(\textbf{x})$ respectively, deviate as much as possible. We introduce a denoising function $D$ as an adversary in the process leading to Equation~\ref{eq:cloakv1-2}.

\begin{equation}
\text{max}_{\textbf{\^{x}}}\mathcal{L}_{total}(F(D(\textbf{\^{x}})),F(\textbf{x}))\text{ s.t. }|D(\textbf{\^{x}})-\textbf{x}|_\infty < \epsilon 
\label{eq:cloakv1-2}
\end{equation}

Note that we apply the denoising step \textit{after} the cloaking step and as such given an initial \^{x}$_0$, $D($\text{\^{x}}$_0) = $\text{ \^{x}}$_0$.

\para{Defense setup.} We follow the defense setup of Case Study 1, where we choose the black-box cloak of UnGANable, Cloak v1, as the protection scheme. We selected SDXL on the best performing prompt C6 as the countermeasure.  We start with 20 images, of which we select 17 images in which we succeeded in our attack (\ie without the countermeasure). We perform this experiment with a perturbation budget of 0.05. When integrating SDXL as a countermeasure, we also integrate the Stable Diffusion Upscaler~\cite{rombach2022high} to ensure that the input is $512 \times 512$. We follow the same strategy as our denoising pipeline for Case Study 1 (Appendix~\ref{appendix:unganable-settings}).

\para{Attack evaluation metrics.} We use the same performance and utility metrics as Case Study 1. (1) \textit{Matching Rate:} Percentage of reconstructed images that match the identity of the corresponding target images, indicating a successful attack. (2) \textit{Utility measures:} We use PSNR, SSIM, and MSE to measure image utility (same interpretation as Case Study 1). 

\begin{table}[!t]
\centering
\small 
\setlength{\tabcolsep}{1.5pt}
\setlength\extrarowheight{2pt}
\caption{The performance and utility of Cloak v1 vs. Cloak v1 with an SDXL countermeasure. The countermeasure fails to add significant protection to images that were unsuccessfully cloaked and harms utility.}
\begin{tabular}{l"c|c|c|c}
                             \textbf{Attacker} & \textbf{\begin{tabular}[c]{@{}c@{}}Matching\\ Rate ↑\end{tabular}} & \textbf{PSNR ↑} & \textbf{SSIM ↑} & \textbf{MSE ↓} \\ \Xhline{1.1pt}
\textbf{UnGANable Cloak v1}  & 100.0\%                     & 33.819          & 0.943           & 0.0004         \\
\textbf{+ SDXL C6 Denoising} & 82.4\%                     & 29.155          & 0.928           & 0.0013        
\end{tabular}
\label{tab:unganable-countermeasure}
\end{table}

\subsection{Countermeasure Study 2: Denoising-aware SIREN}\label{appendix:siren-countermeasure}
\noindent This appendix section outlines our setup and findings for our countermeasure strategy for SIREN.

\para{Countermeasure setup.} 
We follow the defense setup of Case Study 4, where we use the Pokemon~\cite{Pokemon:online} dataset to personalize a SD1.5 model to generate 1000 images.  The one difference is that we fine-tune our own encoders and decoders instead of using their provided checkpoints for the dataset.  We start with their generalized checkpoints as a starting point and fine-tune with the LoRA~\cite{hu2022lora} checkpoint obtained by running personalization on clean images from Case Study 4. From experimentation, 20 epochs is sufficient to create an encoder and decoder that can properly apply the SIREN coating.  

SIREN's total loss, $\mathcal{L}_{total}$, consists of several loss components defined by the training objective in Equation~\ref{eq:sirenloss}.

\begin{equation}
\text{min}_\textbf{}\mathcal{L}_{total} = \lambda_1\mathcal{L}_{learn} + \lambda_2\mathcal{L}_{percept} + \lambda_3(\mathcal{L}_{hc}^+ + \mathcal{L}_{hc}^-) 
\label{eq:sirenloss}
\end{equation}

$\mathcal{L}_{learn}$ represents the learnability loss which ensures that personalized methods are able to learn the coating as being feature relevant.  $\mathcal{L}_{perceptual}$ represents the perceptual loss which ensures that the coating is imperceptible.  $\mathcal{L}_{hc}^+$ and $\mathcal{L}_{hc}^-$ represent hypersphere classification losses of positive and negative samples, respectively, which ensure that the coating on generated images are detectable.
$\lambda_1, \lambda_2,$ and $\lambda_3$ represent hyperparameters to alter the importance of each loss component.

For each epoch where SIREN creates adversarial images through added noise, we integrate our pipeline to denoise those images to serve as an adversary. We specifically choose FLUX and follow the setting from PRC Watermark (Case Study 2).  We also fine-tune an encoder and decoder without the countermeasure for the same 20 epochs for fair comparison.  The same attack configuration, including evaluation metrics, of Case Study 4 are reused for this experiment.  We use our fine-tuned ``no countermeasure'' decoder as the detection scheme for calculating performance.

\begin{table}[!t]
\centering
\small 
\setlength{\tabcolsep}{1.5pt}
\setlength\extrarowheight{2pt}
\caption{The performance and utility of SIREN vs. SIREN with a FLUX countermeasure. Adding the countermeasure reduces the protection to 0.}
\begin{tabular}{l"c|c|c|c|c}

                             \textbf{Attacker} & \textbf{\begin{tabular}[c]{@{}c@{}}TPR@\\ Sign. ↓\end{tabular}} & \textbf{KID ↓} & \textbf{PSNR ↑} & \textbf{SSIM ↑}  & \textbf{LPIPS ↓} \\ \Xhline{1.1pt}
\textbf{SIREN Coating}  & 0.991                     & 0.078          & 36.040           & 0.816 & 0.030        \\
\textbf{+ FLUX} & 0.000                     & 0.069          & 36.866          & 0.827 & 0.029       
\end{tabular}

\label{tab:siren-countermeasure}

\end{table}

\begin{figure}[t]
\centering
\includegraphics[scale=0.17,angle=270]{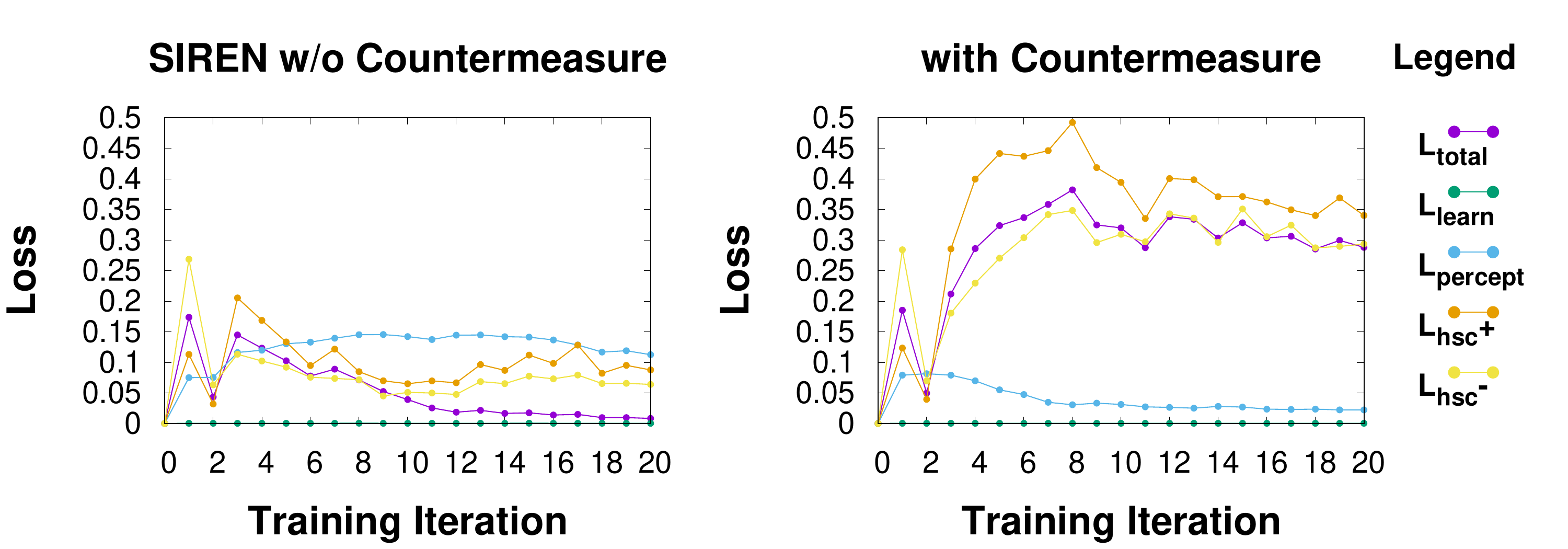}
\caption{Loss curve for SIREN with and without the countermeasure. With the countermeasure, loss values increase and fluctuate (right). Without the countermeasure, the loss properly minimizes resulting in effective SIREN coating (left).}
\label{fig:siren-countermeasure}

\end{figure}

\para{Results.} The results are in Table~\ref{tab:siren-countermeasure}.

\noindent\textbf{\textit{RQ5:}} Similarly to our countermeasure design for UnGANable, applying the countermeasure does not aid the protection.  We find that the TPR@Significance decreases from 0.991, which is close to 1 for ideal cloaking, to an abysmal 0.000.  This suggests that the countermeasure is hindering the process of learning an effective SIREN coating.  This is further supported by examining the loss curves (Figure~\ref{fig:siren-countermeasure}).  The total loss increases over time and hovers between 0.25 and 0.40 after epoch 8.  Without the countermeasure, the total loss properly minimizes as the number of iterations increase. 

Examining the loss components, we find that $\mathcal{L}_{percept}$, $\mathcal{L}_{hc}^+$, and $\mathcal{L}_{hc}^-$ exhibit different behavior for the countermeasure.  $\mathcal{L}_{hc}^+$, and $\mathcal{L}_{hc}^-$ closely follow the total loss curve which indicates that the detectability is not well optimized.  This is a significant flaw of the countermeasure as it removes the data tracing feature of SIREN.  $\mathcal{L}_{perceptual}$ also more properly minimizes for the countermeasure, indicating that applying the countermeasure causes the fine-tuning process to instead optimize the imperceptibility of the coating.  Considering that the default coating provides high quality, this is not a benefit to the countermeasure.  Ultimately, a countermeasure cannot be designed for SIREN as it disrupts the coating process entirely. 

\begin{table}[t]
\centering
\small
\setlength{\tabcolsep}{1.5pt}
\caption{Strength ablation for PRC Watermark. Modifying the strength shows a tradeoff between performance and utility.}
\begin{tabular}{l"c|c|c|c}
                       \textbf{FLUX} & \textbf{TPR@FPR ↓} & \textbf{PSNR ↑} & \textbf{SSIM ↑} & \textbf{KID ↓} \\ \Xhline{1.1pt}
\textbf{Strength 0.05} & 0.982              & \textbf{32.399}          & \textbf{0.894}          & 0.0198         \\
\textbf{Strength 0.15} & 0.258              & 28.042          & 0.775          & \textbf{0.0196}         \\
\textbf{Strength 0.25} & \textbf{0.002}              & 25.944          & 0.703          & 0.0214        
\end{tabular}
\label{tab:prc-ablation-str}
\end{table}

\begin{table}[t]
\centering
\small
\setlength{\tabcolsep}{1.5pt}
\caption{Strength ablation for SIREN. Modifying the strength shows a tradeoff between performance and utility.}
\begin{tabular}{l"c|c|c|c|c}
                       \textbf{FLUX} & \textbf{TPR@Sign. ↓} & \textbf{KID ↓}& \textbf{PSNR ↑} & \textbf{SSIM ↑} &  \textbf{LPIPS ↓} \\ \Xhline{1.1pt}
\textbf{Strength 0.25} & 0.089             & \textbf{0.066} & \textbf{31.748}          & \textbf{0.828}   & \textbf{0.034}       \\
\textbf{Strength 0.35} & 0.016            & 0.071 & 28.882          & 0.787   & 0.050        \\
\textbf{Strength 0.45} & \textbf{0.000}             & 0.079 & 26.188          & 0.731   & 0.075        
\end{tabular}
\label{tab:siren-ablation-str}
\end{table}

\subsection{No-prompt Denoising}\label{appendix:section8-2}

\begin{table}[!t]
\centering
\small
\setlength{\tabcolsep}{1.5pt}
\setlength\extrarowheight{2pt}
\caption{No-prompt denoising results for PRC Watermark. The no-prompt result has a higher TPR@FPR than the with prompt result.}
\begin{tabular}{l"c|c|c|c}
\textbf{Attacker}         & \textbf{TPR@FPR ↓} & \textbf{PSNR ↑} & \textbf{SSIM ↑} & \textbf{KID ↓}  \\ \Xhline{1.1pt}
\textbf{SD1.5}            & 0.862              & 23.334          & 0.612           & 0.0244          \\
\textbf{SDXL}             & \textbf{0.004}     & 24.172          & 0.657           & \textbf{0.0144} \\
\textbf{SD3}              & 0.286              & 25.656 & 0.701  & 0.0187 \\
\rowcolor[gray]{0.85} \textbf{FLUX}             & 0.420     & \textbf{28.478} & \textbf{0.791}  & 0.0199
\end{tabular}
\label{tab:prc-watermark-nonprompt}
\end{table}

\begin{table}[!t]
\centering
\small
\setlength{\tabcolsep}{1.5pt}
\setlength\extrarowheight{2pt}
\caption{No-prompt denoising results for VINE-R. The no-prompt result has a higher TPR@FPR than the with prompt result. }
\begin{tabular}{l"c|c|c|c|c}
\textbf{Attacker}         & \textbf{TPR@FPR ↓} & \textbf{PSNR ↑} & \textbf{SSIM ↑} & \textbf{LPIPS ↓} & \textbf{KID ↓}  \\ \Xhline{1.1pt}
\textbf{SD1.5}            & 0.987             & 22.863          & 0.650           & 0.203 & 0.0002        \\
\textbf{SDXL}             & \textbf{0.774}     & 23.269         & 0.682           & 0.204 & 0.0008 \\
\textbf{SD3}              & 0.854              & 23.911 & 0.691  & 0.179 & 0.0003 \\
\rowcolor[gray]{0.85} \textbf{FLUX}             & 0.956     & \textbf{27.415} & \textbf{0.797}  & \textbf{0.132} & \textbf{0.0001}
\end{tabular}
\label{tab:vine-nonprompt}
\end{table}

\begin{table}[!t]
\centering
\small
\setlength{\tabcolsep}{1.5pt}
\setlength\extrarowheight{2pt}
\caption{No-prompt denoising results for SIREN. The no-prompt result has a higher TPR@Significance than the with prompt result. }
\begin{tabular}{l"c|c|c|c|c}
                      \textbf{Attacker} & \textbf{TPR@Sign. ↓} & \textbf{KID ↓} & \textbf{PSNR ↑} & \textbf{SSIM ↑}  & \textbf{LPIPS ↓} \\ \Xhline{1.1pt}
\textbf{SD1.5}        & 0.120              & 0.091         & 22.401  & 0.812 & 0.121           \\
\textbf{SDXL}         & \textbf{0.000}     & 0.101          & 22.541           & 0.745          & 0.119            \\
\textbf{SD3}          & 0.015              & 0.096 & 23.179           & 0.636          & 0.111   \\
\rowcolor[gray]{0.85} \textbf{FLUX}         & 0.472     & \textbf{0.079} & \textbf{29.243}  & \textbf{0.827} & \textbf{0.054}  
\end{tabular}
\label{tab:siren-noprompt}
\end{table}

\begin{table}[!t]
\centering
\small
\setlength{\tabcolsep}{1.5pt}
\caption{Supervised results for UnGANable.  Supervised SDXL does not outperform SD3.}
\begin{tabular}{l"c|c|c|c}
    & \multicolumn{4}{c}{\textbf{$\epsilon = $ 0.06}}\\ \cline{2-5}
    \textbf{Attacker} & \textbf{\begin{tabular}[c]{@{}c@{}}Matching\\ Rate ↑\end{tabular}} & \textbf{PSNR ↑}   & \textbf{SSIM ↑}  & \textbf{MSE ↓}    \\ \Xhline{1.1pt}
\rowcolor[gray]{0.85} \textbf{SD3}             & \textbf{77.78\%}                                                  & 31.488          & 0.937          & \textbf{0.0007} \\
\textbf{FLUX}            & 76.07\%                                                           & \textbf{31.552} & \textbf{0.941} & \textbf{0.0007} \\

\textbf{SDXL}            & 63.68\%                                                           & 30.726          & 0.925          & 0.0010 \\ \Xhline{1.1pt}
\textbf{Supervised SDXL} & 69.66\%                                                          & 25.302          & 0.873          & 0.0034         
\end{tabular}

\label{tab:unganable-supervised}
\end{table}

\begin{table*}[!t]
\centering
\small
\setlength{\tabcolsep}{1.5pt}
\caption{Prompt ablation for UnGANable.  C2 and C6 are the best performing prompts.}
\begin{tabular}{l"c|c|c|c"c|c|c|c"c|c|c|c}
\textbf{}                & \multicolumn{4}{c"}{\textbf{$\epsilon = $ 0.05}}                                                                                & \multicolumn{4}{c"}{\textbf{$\epsilon = $ 0.06}}                                                                                & \multicolumn{4}{c}{\textbf{$\epsilon = $ 0.07}}                                                                                \\ \cline{2-13}
\textbf{SD3}        & \textbf{\begin{tabular}[c]{@{}c@{}}Matching\\ Rate ↑\end{tabular}} & \textbf{PSNR ↑}   & \textbf{SSIM ↑}  & \textbf{MSE ↓}    & \textbf{\begin{tabular}[c]{@{}c@{}}Matching\\ Rate ↑\end{tabular}} & \textbf{PSNR ↑}   & \textbf{SSIM ↑}  & \textbf{MSE ↓}    & \textbf{\begin{tabular}[c]{@{}c@{}}Matching\\ Rate ↑\end{tabular}} & \textbf{PSNR ↑}   & \textbf{SSIM ↑}  & \textbf{MSE ↓}    \\ \Xhline{1.1pt}
\textbf{Prompt C1}             & 68.21\%                                                 & 31.672 & 0.939 & \textbf{0.0007} & 67.09\%                                                  & 31.204 & 0.931 & 0.0008 & 61.96\%                                                 & 30.679 & 0.921 & 0.0009 \\
\textbf{Prompt C2}             & \textbf{77.44\%}                                                 & \textbf{31.923} & \textbf{0.943} & \textbf{0.0007} & 72.65\%                                                  & \textbf{31.495} & \textbf{0.937} & \textbf{0.0007} & 68.48\%                                                & \textbf{31.044} & \textbf{0.929} & \textbf{0.0008} \\
\textbf{Prompt C3}             & 73.33\%                                                 & 31.632 & 0.939 & \textbf{0.0007} & 65.81\%                                                  & 31.108 & 0.930 & 0.0010 & 63.04\%                                                 & 30.579 & 0.920 & 0.0013\\
\textbf{Prompt C4}             & \textbf{77.44\%}                                                 & 31.820 & 0.942 & \textbf{0.0007} & 67.09\%                                                  & 31.351 & 0.935 & 0.0008 & 67.39\%                                                 & 30.870 & 0.926 & \textbf{0.0008}\\
\textbf{Prompt C5}             & 70.26\%                                                 & 31.671 & 0.939 & \textbf{0.0007} & 63.68\%                                                  & 31.193 & 0.931 & 0.0008 & 61.96\%                                                 & 30.640 & 0.921 & 0.0009\\
\rowcolor[gray]{0.85} \textbf{Prompt C6}             & 75.38\%                                                  & 31.908 & \textbf{0.943} & \textbf{0.0007} & \textbf{77.78\%}                                                  & 31.488 & \textbf{0.937} & \textbf{0.0007} & \textbf{71.01\%}                                                  & 31.029 & \textbf{0.929} & \textbf{0.0008} \\
\textbf{Prompt C7}             & 71.28\%                                                 & 31.634 & 0.939 & \textbf{0.0007} & 66.24\%                                                  & 31.108 & 0.930 & 0.0009 & 62.68\%                                                 & 30.652 & 0.921 & 0.0009 \\
\textbf{Prompt C8}             & 73.33\%                                                 & 31.803 & 0.942 & \textbf{0.0007} & 68.38\%                                                  & 31.342 & 0.934 & 0.0008 & 65.22\%                                                 & 30.864 & 0.926 & \textbf{0.0008} \\ \Xhline{1.1pt}
\textbf{Average}             & 73.33\%                                                 & 31.758 & 0.941 & 0.0007 & 68.59\%                                                  & 31.286 & 0.933 & 0.0008 & 65.22\%                                                 & 30.795 & 0.924 & 0.0009 \\
\textbf{Standard Deviation}             & 3.33\%                                                 & 0.121 & 0.002 & 0.0000 & 4.52\%                                                  & 0.156 & 0.003 & 0.0001 & 3.41\%                                                 & 0.182 & 0.004 & 0.0002 \\
\end{tabular}
\label{tab:prc-ablation-prompt}
\end{table*}

\begin{table*}[t]
\centering
\small
\setlength{\tabcolsep}{1.5pt}
\setlength\extrarowheight{2pt}
\caption{No-prompt denoising results for UnGANable. The no-prompt result for SD3 has a lower Matching Rate than the with prompt result. }
\begin{tabular}{l"c|c|c|c"c|c|c|c"c|c|c|c}
\textbf{}                & \multicolumn{4}{c"}{\textbf{$\epsilon = $ 0.05}}                                                                                & \multicolumn{4}{c"}{\textbf{$\epsilon = $ 0.06}}                                                                                & \multicolumn{4}{c}{\textbf{$\epsilon = $ 0.07}}                                                                                \\ \cline{2-13}
\textbf{Attacker}        & \textbf{\begin{tabular}[c]{@{}c@{}}Matching\\ Rate  ↑\end{tabular}} & \textbf{PSNR ↑}   & \textbf{SSIM ↑}  & \textbf{MSE ↓}    & \textbf{\begin{tabular}[c]{@{}c@{}}Matching\\ Rate ↑\end{tabular}} & \textbf{PSNR ↑}   & \textbf{SSIM ↑}  & \textbf{MSE ↓}    & \textbf{\begin{tabular}[c]{@{}c@{}}Matching\\ Rate  ↑\end{tabular}} & \textbf{PSNR ↑}   & \textbf{SSIM ↑}  & \textbf{MSE ↓}    \\ \Xhline{1.1pt}
\textbf{SD1.5}           & 68.21\%                                                          & 31.314         & 0.936         &  0.0008         & 66.24\%                                                           & 30.870          & 0.928          & 0.0009          & 64.13\%                                                            & 30.402         & 0.919          & 0.0009         \\
\textbf{SDXL}            & 75.38\%                                                           & \textbf{32.513}          & \textbf{0.948}          & \textbf{0.0006} & 74.79\%                                                           & \textbf{32.064}          & \textbf{0.941}           & \textbf{0.0006} & \textbf{72.46\%}                                                           & \textbf{31.580}           & \textbf{0.935}            & \textbf{0.0007} \\
\textbf{SD3}             & 73.85\%                                                  & 31.909 & 0.943 & 0.0007 & 70.94\%                                                 & 31.432 & 0.935 & \textbf{0.0007} & 68.12\%                                                  & 30.940 & 0.927 & 0.0008 \\
\rowcolor[gray]{0.85} \textbf{FLUX}            & \textbf{76.92\%}                                                  & 32.150 & 0.947 & \textbf{0.0006} & \textbf{76.07\%}                                                  & 31.708 & 0.940 & \textbf{0.0007} & 72.10\%                                                  & 31.238 & 0.932 & 0.0008
\end{tabular}
\label{tab:ungannable-nonprompt}
\end{table*}

\para{Results.} Our no-prompt results can be found in Tables~\ref{tab:prc-watermark-nonprompt},~\ref{tab:vine-nonprompt},~\ref{tab:siren-noprompt}, and~\ref{tab:ungannable-nonprompt}. For UnGANable, the Matching Rate at the best no-prompt setting is 76.92\% which is worse than the best with prompt results of 77.78\%.  The best models for UNGANable, SD3 and FLUX, obtain similar or worse performance overall to their ``with prompt'' variants.  SDXL no-prompt improves to a competitive 75.38\% Matching Rate and provides the best utility of 32.513 PSNR, 0.948 SSIM, and 0.0006 MSE, but it still fails to outperform ``with prompt'' SD3 and FLUX. 

Performance metrics for PRC Watermark are worse overall for no-prompt.  This is especially true for FLUX in which a worse TPR@FPR of 0.420 is obtained.  VINE's performance for no-prompt is similarly worse overall for all models in our pipeline.  This includes FLUX and SDXL in which we obtain 0.956 and 0.774, respectively.  For SIREN, we obtain a significantly higher TPR of 0.472 for FLUX (``with prompt'' result is 0.016) and similar performance for the other three models.

The decrease in performance ultimately outweighs the increase in utility and also shows that setting a prompt can help improve results. As a result, the ``with prompt'' setting is preferred to answer \textbf{\textit{RQ1}} through \textbf{\textit{RQ3}} for these case studies.

\subsection{Supervised Denoising}\label{appendix:section8-3}

\noindent The results for supervised denoising are shown in Table~\ref{tab:unganable-supervised}. The backbone of this process is Instruction-tuned Stable Diffusion~\cite{paul2023instruction}, where a Stable Diffusion model is fine-tuned on a paired dataset of original and protected images with an edit instruction linking the pairs. Instruction-tuning was introduced through FLAN~\cite{wei2021finetuned} and popularized by InstructPix2Pix~\cite{brooks2023instructpix2pix}.  The goal is to learn the cloaking patterns of UnGANable and denoise them for further improvement.

We create a paired dataset of 5000 generated images (following Case Study 1's generation) and set the edit instruction to be the positive prompt of C6. We use this dataset to finetune a ``diffusers/sdxl-instructpix2pix-768'' model~\cite{supervised:online} to remove the cloak and subsequently use the model for denoising. We follow the default configuration of Instruction-tuned Stable Diffusion, with a learning rate of $5e^{-5}$ and training steps of 15,000 for the fine-tune process. We follow the same attack configuration and evaluation metrics as Case Study 1.


\end{document}